\documentclass{article}
\usepackage{natbib}

\usepackage{amsfonts}
\usepackage{bbding}
\usepackage{xcolor}
\usepackage{float}
    \usepackage[preprint,nonatbib]{neurips_2025}

\usepackage[utf8]{inputenc} 
\usepackage[T1]{fontenc}    
\usepackage{hyperref}       
\usepackage{url}            
\usepackage{booktabs}       
\usepackage{amsfonts}       
\usepackage{nicefrac}       
\usepackage{microtype}      
\usepackage{xcolor}         
\usepackage{tikz}
\usepackage{makecell}
\usepackage{array}
\usepackage{subcaption}
\usepackage{longtable}
\usepackage{multirow}

\usepackage{bm}
\usetikzlibrary{arrows.meta, positioning, shapes.geometric}

\title{VUDG: A Dataset for Video Understanding Domain Generalization}

\author{%
Ziyi Wang$^{1}$, Zhi Gao$^{1}$\thanks{~Corresponding author.}, ~Boxuan Yu$^{1}$, Zirui Dai$^{1}$, Yuxiang Song$^{1}$, \\ \textbf{Qingyuan Lu}$^{1}$, \textbf{Jin Chen}$^{1}$, \textbf{Xinxiao Wu}$^{1,2}$ \\
\small $^1$Beijing Key Laboratory of Intelligent Information Technology, \\ \small School of Computer Science \& Technology, Beijing Institute of Technology \\
\small \textsuperscript{2}Guangdong Laboratory of Machine Perception and Intelligent Computing, Shenzhen MSU-BIT University \\
\small \hyperlink{https://VUDG-Video.github.io}{\texttt{https://VUDG-Video.github.io}}
}

\begin{document}

\maketitle

\vskip -0.15in
\begin{abstract}
\vskip -0.1in
Video understanding has made remarkable progress in recent years, largely driven by advances in deep models and the availability of large-scale annotated datasets. 
However, existing works typically ignore the inherent domain shifts encountered in real-world video applications, leaving domain generalization (DG) in video understanding underexplored.
Hence, we propose \textbf{V}ideo \textbf{U}nderstanding \textbf{D}omain \textbf{G}eneralization (\textbf{VUDG}), a novel dataset designed specifically for evaluating the DG performance in video understanding.
VUDG contains videos from 11 distinct domains that cover three types of domain shifts, and maintains semantic similarity across different domains to ensure fair and meaningful evaluation. We propose a multi-expert progressive annotation framework to annotate each video with both multiple-choice and open-ended question-answer pairs. 
Extensive experiments on 9 representative large video-language models (LVLMs) and several traditional video question answering methods show that most models (including state-of-the-art LVLMs) suffer performance degradation under domain shifts. 
These results highlight the challenges posed by VUDG and the difference in the robustness of current models to data distribution shifts. We believe VUDG provides a valuable resource for prompting future research in domain generalization video understanding.

\end{abstract}

\section{Introduction}
\vskip -0.1in

Video understanding has achieved remarkable progress, driven by advances in deep models and the availability of large-scale annotated datasets. Existing methods have shown outstanding performance in diverse tasks such as action recognition~\cite{chen2022mm, qing2023mar, liu2025mnv3}, video captioning~\cite{lin2022swinbert, seo2022end, gu2023text, kavitha2024automatic}, and video question answering (VideoQA)~\cite{li2022invariant, li2022representation, min2024morevqa}. However, most existing methods typically ignore the inherent domain shifts encountered in real-world video applications, which assume the same distribution between training and testing data. This leads to the degradation of model performance, and the generalization of video understanding models is limited. 
This problem could be formulated as  
the task of domain generalization (DG)~\cite{lin2023diversifying, papadakis2024multi,chen2023meta} in video understanding, where the model trained on training data (source domain) is expected to perform well on unseen testing data (target domain) with different distributions. 
In recent years, this task has received increasing attention with the urgent demand to apply Large Video Language Models (LVLMs) to downstream video understanding tasks, while the performance of LVLMs may degrade due to the distribution differences between the training and unseen testing data. Hence, it is essential and valuable to study domain generalization in video understanding.
Although several benchmarks~\cite{jang2017tgif,li2024mvbench,li2024videovista,fu2024video} have started to focus on video understanding across multiple domains with different distributions, they are not suitable to evaluate the DG performance, as the semantic spaces across domains are different, and the performances of models will be affected by not only the domain shift but also the semantic space differences. Hence, the generalization of the model may not be evaluated well.
To address this issue, we propose Video Understanding Domain Generalization (VUDG), a dataset specifically designed to evaluate the performance of domain generalization in video understanding. 
VUDG contains 11 domains with video data collected from multiple open-source datasets~\cite{wang2019vatex, ugai2024multimodal, wang2024internvid, chen2024sharegpt4video} in order to include as diverse a distribution as possible.
These domains exhibit variations in visual styles (\emph{e.g.}, cartoon, game, movie/TV show, virtual environment), viewpoints (\emph{e.g.}, egocentric, surveillance, shaky), and environmental conditions (\emph{e.g.}, foggy, night, rainy, snowy), making VUDG a comprehensive generalization evaluation in video understanding.
Note that we collect both the training and testing data for each domain to avoid the potential data leakage issue in evaluating LVLMs, and they are collected from distinct data sources.


\begin{figure}
    \centering
    \includegraphics[width=1\linewidth]{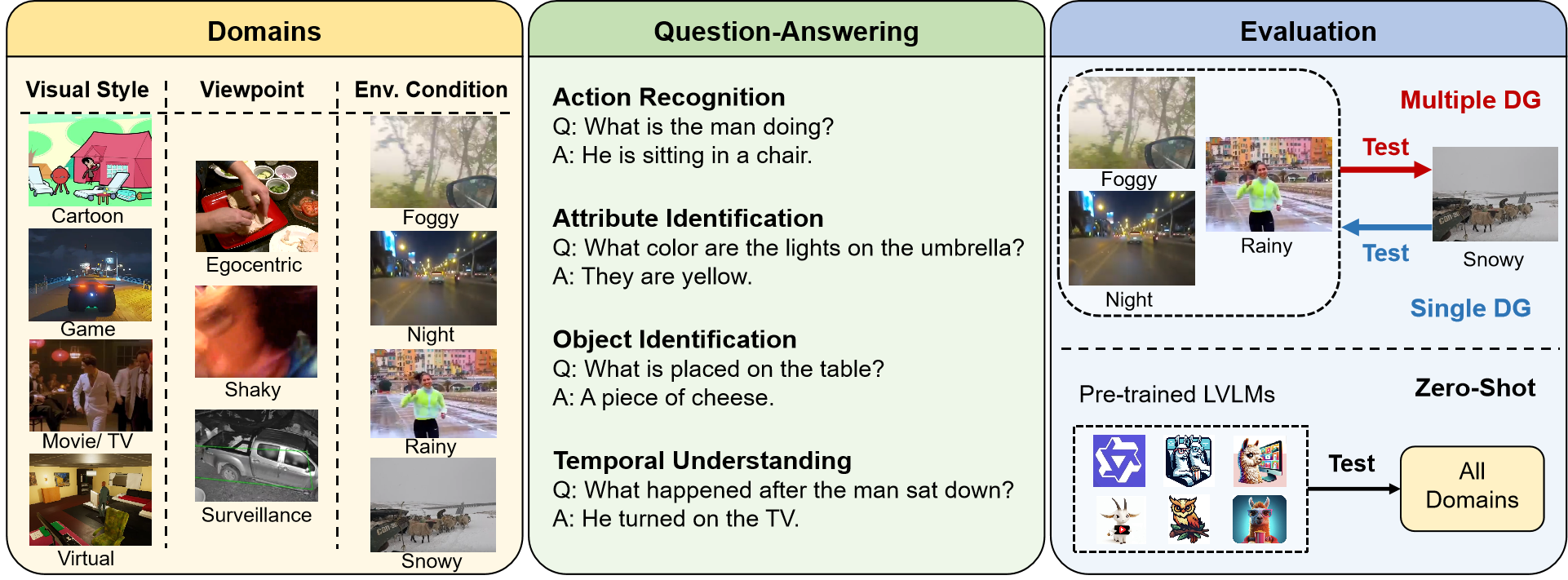}
    \caption{Overview of the proposed VUDG dataset.}
    \vspace{-15pt}
    \label{fig:teaser}
\end{figure}

We propose a progressive multi-expert annotation framework for video annotation.
Firstly, we select 11 domains and summarize a list of human daily activity scenes as the semantic space by brainstorming, and then employ Qwen2.5-VL-7B~\cite{bai2025qwen2} to select videos from existing video source datasets.
Secondly, each video is annotated with open-ended question-answer (QA) pairs using Gemini-2.5-Flash, as it is fast and cost-effective. These open-ended QA pairs are then used to generate multiple-choice options by DeepSeek-v3.
Subsequently, these preliminary annotations are checked and filtered using Gemini-2.5-Pro for quality control. Finally, we ask humans to review and correct the uncertain or erroneous samples identified in the previous step to obtain high-quality QA pairs. For multiple-choice QA, we directly calculate the accuracy, and for open-ended QA pairs, we use DeepSeek-v3 to evaluate the answers.
Here, we use different large models in different stages, alleviating the hallucination and bias problems caused by using a fixed model.
Our dataset is designed to support multiple domain generalization and single domain generalization tasks, enabling rigorous evaluation of models' domain generalization ability when adapting to downstream tasks.
In addition, we can also measure the zero-shot generalization performance of video understanding methods using only the testing data of each domain.

We evaluate several representative VideoQA models and LVLMs under the multiple domain generalization and single domain generalization settings. We find that these models have limited generalization capabilities to unseen domains, resulting in unsatisfactory performance. This emphasizes the importance of studying domain generalization in video understanding.
Furthermore, we evaluate 9 LVLMs on the zero-shot generalization setting, and the results show that most models do not achieve satisfactory performance.
These findings underscore the difficulty of achieving robust generalization in video understanding.
These experiments suggest that VUDG can serve as a reliable dataset for domain generalization in video understanding and motivate future work on developing more generalizable multimodal models.

In summary, our key contributions are as follows:

\begin{itemize}
\item We introduce VUDG, the first dataset for evaluating domain generalization in video understanding, which contains 11 different domains, including videos with varying visual styles, viewpoints, and environmental conditions.
\item We propose a progressive multi-expert annotation framework that sequentially combines multiple expert models and human experts for annotation generation and filtering, effectively obtaining high-quality open-ended and multiple-choice QA pairs.
\item We conduct comprehensive experiments on 9 state-of-the-art LVLMs and several VideoQA baselines, revealing the challenge of VUDG on large gaps across domains and highlighting the limitations of current models under distribution shifts.
\end{itemize}

\section{Related Work}

\subsection{Video Understanding Benchmarks}

The development of video understanding datasets has fueled recent advancements in video understanding. Datasets such as ActivityNet~\cite{caba2015activitynet} and Kinetics~\cite{kay2017kinetics} have been pivotal in action recognition tasks, providing millions of labeled video clips across diverse activities. Later, Charades~\cite{sigurdsson2016hollywood}, TVQA~\cite{lei2018tvqa}, MSVD-QA~\cite{xu2017video}, and MSR-VTT-QA~\cite{xu2016msr} expand the scope of video understanding to include VideoQA tasks, introducing videos paired with textual questions that require reasoning about the video content. 
However, these datasets ignore scenarios where the distributions of the training and testing data are inconsistent, leaving the problem of domain generalization underexplored.
Although some recent datasets (\emph{e.g.}, VideoVista~\cite{li2024videovista}, Video-MME~\cite{fu2024video}) include videos from multiple categories such as HowTo, Film, and Cartoon, the semantic disparities between these categories are often too large. Therefore, they are not ideal for isolating the effects of domain shifts, making it difficult to fairly evaluate a model’s generalization ability across domains in downstream tasks.

\subsection{Domain Generalization in Video Tasks}

Previous works~\cite{wang2024domain,zhang2023video,lin2023diversifying,yao2021videodg} have explored domain generalization (DG) in video tasks to enhance robustness under unseen distributions. VideoDG~\cite{yao2021videodg} introduces an adversarial pyramid network and constructs three DG settings based on different dataset sources, different action consequences, and different camera viewpoints for video classification generalization. Ani‑GIFs~\cite{majumdar2022ani} presents the first synthetic DG dataset using animated GIFs and real videos to study domain shift in action recognition. ARGO1M~\cite{plizzari2023can} samples egocentric clips from Ego4D~\cite{grauman2022ego4d} across diverse scenarios and locations to evaluate cross-context generalization. MDVAD~\cite{flaborea2023multimodal} aggregates six surveillance video datasets to benchmark anomaly detection under environment and camera shifts. 
In contrast to these datasets that focus on domain generalization in video classification, anomaly detection, or action recognition, our dataset is tailored for video understanding, which poses richer visual reasoning challenges and aligns closely with the rapid development of LVLMs.

\begin{table}[htbp]
    \centering
    \caption{The comparison of existing video understanding datasets involves several key aspects: total number of videos
(Videos) and video clips (Clips), number of QA pairs (QA Pairs), annotation method (Anno., where M/A indicates manual/automatic), whether the videos contain diverse domains (Dom.), and whether the semantic spaces across domains are similar (Sem.).}
    \scalebox{0.7}{\begin{tabular}{l|c|c|c|c|c|c}
    \toprule
        Datasets & Videos & Clips & QA Pairs & Anno. & Dom. & Sem. \\
    \midrule
         MSRVTT-QA~\cite{xu2017video} &2,990 &2,990  &72,821 &A &\textcolor{red}{\XSolidBrush} & \textbf{--} \\
        MSVD-QA~\cite{xu2017video} &504 &504  &13,157 &A &\textcolor{red}{\XSolidBrush} & \textbf{--} \\
        ActivityNet-QA~\cite{yu2019activitynet} &800 &800  & 8,000 & M & \textcolor{red}{\XSolidBrush} & \textbf{--} \\
        EgoSchema~\cite{mangalam2023egoschema} &5,063 &5,063 &5,063 &A\&M &\textcolor{red}{\XSolidBrush} & \textbf{--} \\
        TGIF-QA~\cite{jang2017tgif} &9,575 & 9,575 & 8,506 & A\&M &\textcolor{blue}{\checkmark} &\textcolor{red}{\XSolidBrush} \\
        
         MVBench~\cite{li2024mvbench} &3,641 &3,641  &4,000 &A &\textcolor{blue}{\checkmark} &\textcolor{red}{\XSolidBrush} \\
         Video-Bench~\cite{ning2023video} &5,917 &5,917 &17,036 &A\&M &\textcolor{blue}{\checkmark} &\textcolor{red}{\XSolidBrush} \\
         TempCompass~\cite{liu2024tempcompass} &410 &500 &7,540 &A\&M &\textcolor{blue}{\checkmark} &\textcolor{red}{\XSolidBrush} \\
        Video-MME~\cite{fu2024video} &900 &900 &2,700 &M &\textcolor{blue}{\checkmark} &\textcolor{red}{\XSolidBrush} \\
        VideoVista~\cite{li2024videovista} &894 &3,402 &3,402 &A &\textcolor{blue}{\checkmark} &\textcolor{red}{\XSolidBrush} \\
        \midrule
        VUDG (Ours) & 7,899 & 7,899 & 36,388 & A\&M &\textcolor{blue}{\checkmark} &\textcolor{blue}{\checkmark} \\
    \bottomrule
    \end{tabular}}
    \label{tab:my_label}
\end{table}


\label{gen_inst}

\section{VUDG Dataset}
\label{headings}

\begin{figure}[ht]
    \centering
    \includegraphics[width=0.9\textwidth]{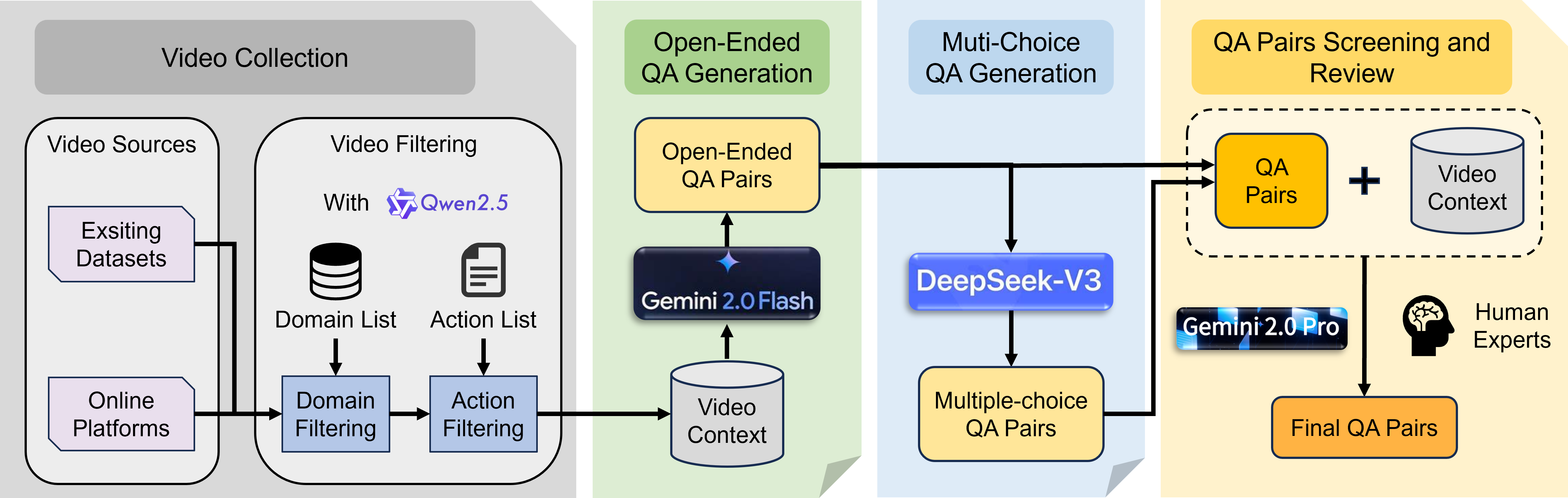}
    \caption{The pipeline diagram of the proposed multi-expert annotation framework.}
    \label{fig:QA_Pipeline}
\end{figure}

We introduce the \textbf{VUDG} dataset that contains 11 distinct domains, including videos with different visual styles, viewpoints, and environmental conditions. To ensure high-quality data, we propose a progressive multi-expert annotation framework that leverages multiple large models, followed by human review for question-answer pairs generation and filtering. Importantly, we incorporate different large models in the generation and verification stages to mitigate the bias stemming from a fixed model that tends to validate its own outputs, thereby improving the diversity, objectivity, and robustness of the collected QA pairs. The annotation pipeline consists of four key stages: \textbf{(1) Video Collection}, \textbf{(2) Open-Ended QA Pairs Generation}, \textbf{(3) Multiple-Choice QA Pairs Generation}, and \textbf{(4) QA Pairs Screening and Review}. The overall workflow is illustrated in Figure~\ref{fig:QA_Pipeline}.

\subsection{Video Collection}
We define 11 domains for video collection, including cartoon, game, movie/TV show, virtual environment, egocentric, surveillance, shaky, foggy, night, rainy, and snowy. Then, we collect videos from various data sources based on the domain name. We employ Qwen2.5-VL-7B~\cite{bai2025qwen2} to filter out irrelevant videos that do not belong to the predefined domains.
To ensure semantic similarity across domains, we manually define a list of daily human activity scenes (\emph{e.g.}, reading books or documents, riding a bicycle, feeding a pet \emph{etc.}) and utilize Qwen2.5-VL-7B to select videos belonging to this predefined activity list.
The activity list and the prompts that are used to select videos are detailed in the Appendix~\ref{prompt_data_collection}.

To eliminate potential data leakage (since LVLMs are pre-trained on a large amount of data), we create training and testing sets for each domain and ensure a clear separation between them by collecting them from different data sources.

\paragraph{Training Set:}  
The training set is constructed exclusively from the training sets of existing open-source datasets, including \textit{InternVid}~\cite{wang2024internvid}, \textit{ShareGPT4Video}~\cite{chen2024sharegpt4video}, \textit{VideoInstruct100K}~\cite{maaz2024videochatgptdetailedvideounderstanding}, and
\textit{MMDL}~\cite{ugai2024multimodal}. We have checked that no data from these sources overlaps with existing benchmarks (used as the testing set), ensuring strict separation between the training and testing data in VUDG.

\paragraph{Testing Set:}  
The testing set is primarily derived from the testing sets of existing open-source video datasets, benchmarks, and videos crawled from online platforms. 
Specifically, we use the test splits of \textit{VATEX}~\cite{wang2019vatex}, \textit{ActivityNet}~\cite{caba2015activitynet}, and 
\textit{MMDL}~\cite{ugai2024multimodal}. Furthermore, we leverage diverse user-generated content from online video platforms such as \textit{YouTube}, \textit{Douyin}, and \textit{Bilibili}. These platforms host rich and varied videos from different domains, enabling us to collect a broad spectrum of videos. 

All subsequent processes, such as QA pairs generation and filtering, are applied uniformly to all the collected videos, without further distinguishing between the training set and the testing set.





\subsection{Question and Answer Generation}

The question and answer generation process comprises 3 steps: open-ended QA pairs generation, multiple-choice QA pairs generation, and QA pairs screening and review.
\begin{figure}[htbp]

    \centering
    \includegraphics[width=0.9\textwidth]{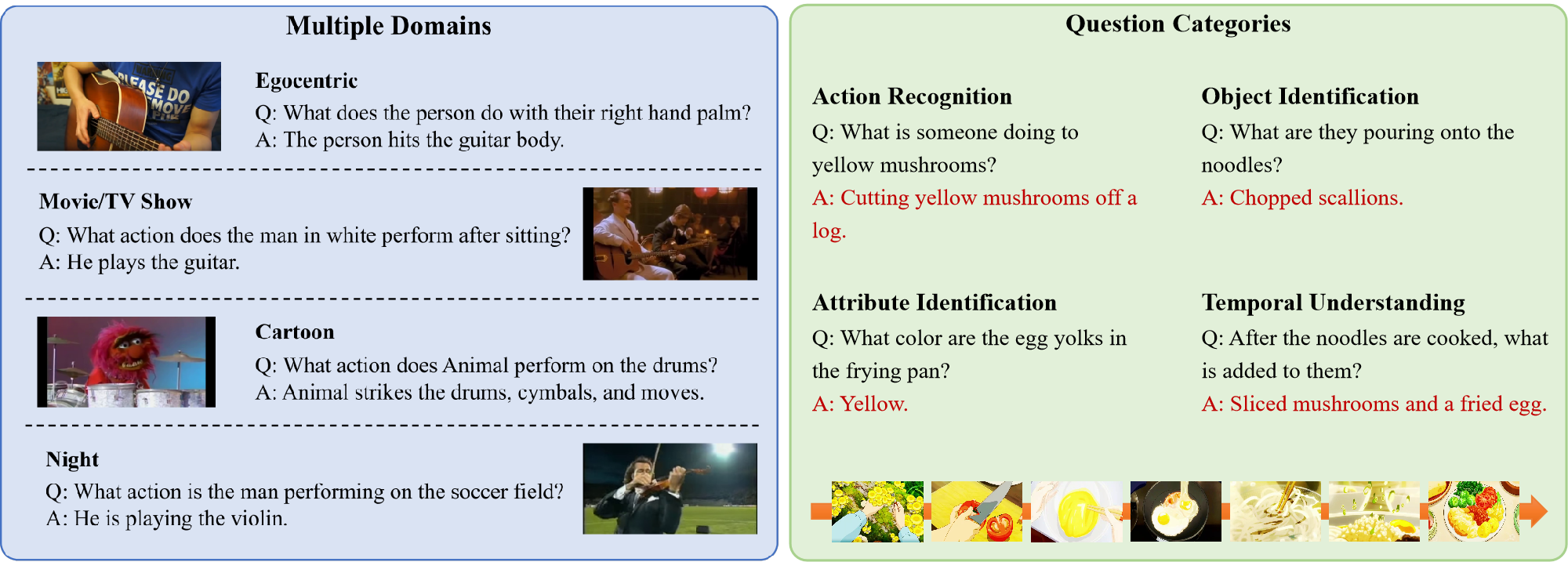}
    \caption{Overview of various domains and question types in our VUDG dataset.}
    \vspace{-5pt}
    \label{fig:QAType}
\end{figure}

\paragraph{Design of Question Category:} 
The generated QA pairs are expected to enable a thorough and comprehensive evaluation of video understanding. Figure~\ref{fig:QAType} illustrates four types of questions: (1) \textbf{Action Recognition}, which focuses on identifying action categories, requiring the model to accurately recognize the action occurring at a specific time point in the video; (2) \textbf{Attribute Identification}, which assesses the model’s ability to perceive visual attributes such as color, shape, and position of simple objects; (3) \textbf{Object Identification}, which tests the model's ability to recognize specific objects; (4) \textbf{Temporal Understanding}, which involves the temporal ordering of actions and requires the model to accurately identify the sequence of events, \emph{i.e.,} the event that occurs before or after a specific action.






\paragraph{Open-Ended QA Pairs Generation:}
To generate open-ended QA pairs, we first leverage Gemini-2.5-Flash, a fast and cost-effective multimodal model with strong video understanding capabilities. This model is used to generate initial questions and open-ended answers for each video. We design two distinct prompts for question categories (1)–(3) and question category (4), respectively, since they focus on different information cues.
Detailed examples of the two prompts can be found in Appendix~\ref{prompt_openend}. For each video, we generate one question for each of the question categories (1)–(3) and two questions for the question categories (4).

\paragraph{Multiple-Choice QA Pairs Generation:}
For the generation of multiple-choice QA Pairs, we utilize DeepSeek-v3 to generate five plausible but incorrect options for each open-ended QA pair. These options are conditioned on the original question and the correct answer. Afterward, the options are randomized to ensure a balanced distribution of all six choices. Examples of prompts for multiple-choice QA generation can be found in Appendix~\ref{prompt_multichoice}.

\paragraph{QA Pairs Screening and Review:}
Despite the structured pipeline used for QA pair generation, issues such as ambiguous phrasing, semantically overlapping options, and factual inaccuracies in open-ended answers may still arise, potentially compromising the quality of the generated pairs. To address these challenges, we introduce a hybrid screening process that integrates both automated model-based evaluation and human-expert review. First, we employ Gemini-2.5-Pro, a more advanced multimodal model, to perform a thorough review of each QA pair with access to the original video context. This model classifies each QA pair into one of three categories: (1) correct QA pairs, (2) partially flawed answers with fixable issues, and (3) invalid questions. These automated classifications serve as the basis for further manual inspection. Then, human experts revise or remove QA pairs that are flagged as problematic ((2) or (3)) by Gemini-2.5-Pro, ensuring that the final dataset maintains high standards of clarity and accuracy. Detailed prompts for Gemini-2.5-Pro can be found in Appendix~\ref{QA_review}.

\subsection{Statistics}

\begin{figure*}[t]
\centering
\begin{subfigure}[t]{0.35\linewidth}
    \includegraphics[width=\linewidth]{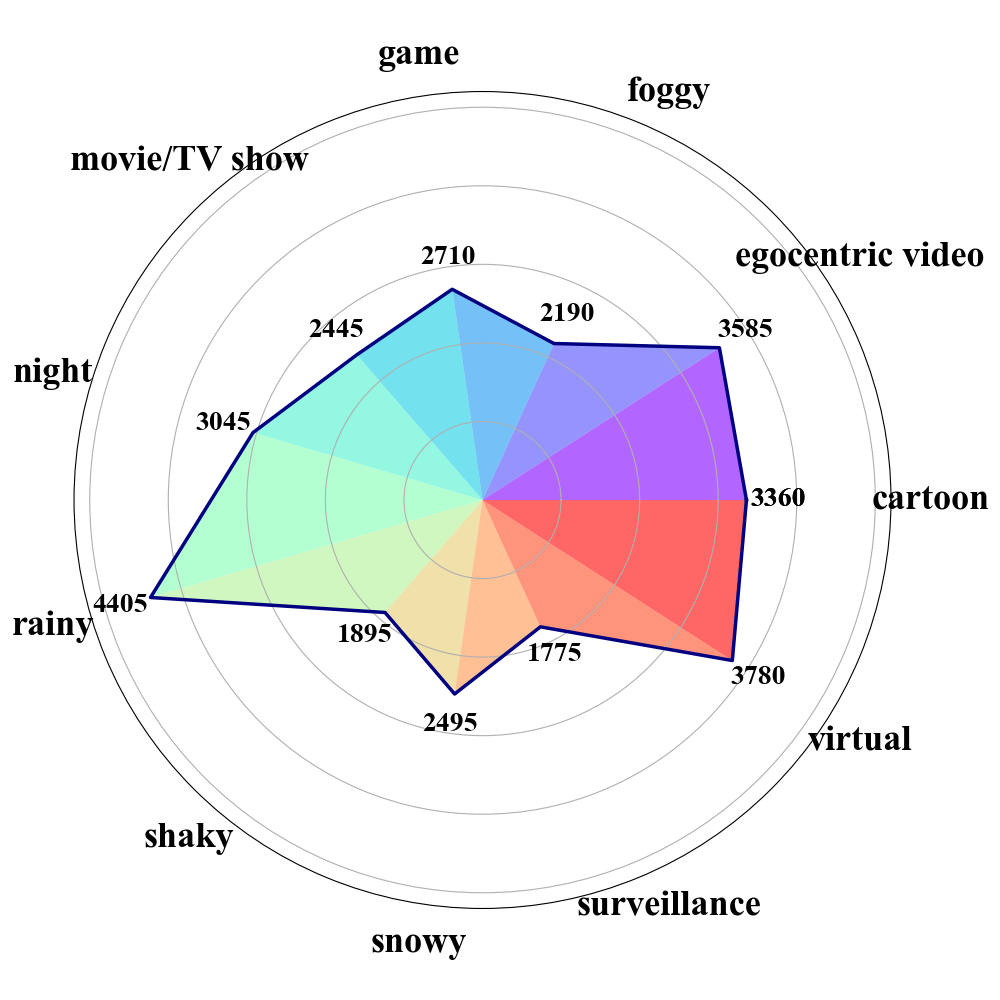}
    \caption{VUDG Training Set}
    \label{fig:vudg_domain_train}
\end{subfigure}
\hspace{40pt}
\begin{subfigure}[t]{0.35\linewidth}
    \includegraphics[width=\linewidth]{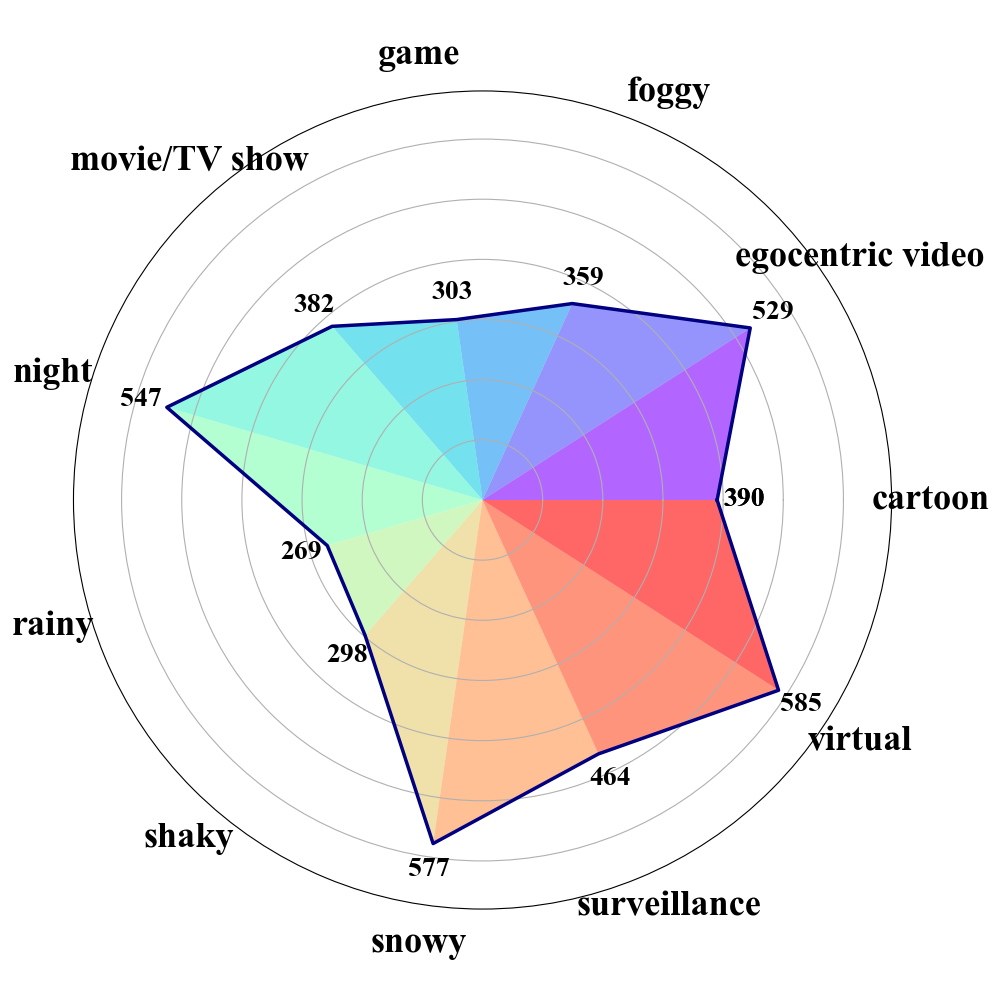}
    \caption{VUDG Testing Set}
    \label{fig:vudg_domain_test}
\end{subfigure}
\caption{Statistics showing the domain and question category distributions of QA pairs in \textbf{(a)} VUDG training set and \textbf{(b)} VUDG testing set.}
\label{fig:vudg_domain_distribution}
\end{figure*}



\begin{figure*}[t]
\centering

\begin{subfigure}[t]{0.3\linewidth}
    \includegraphics[width=\linewidth]{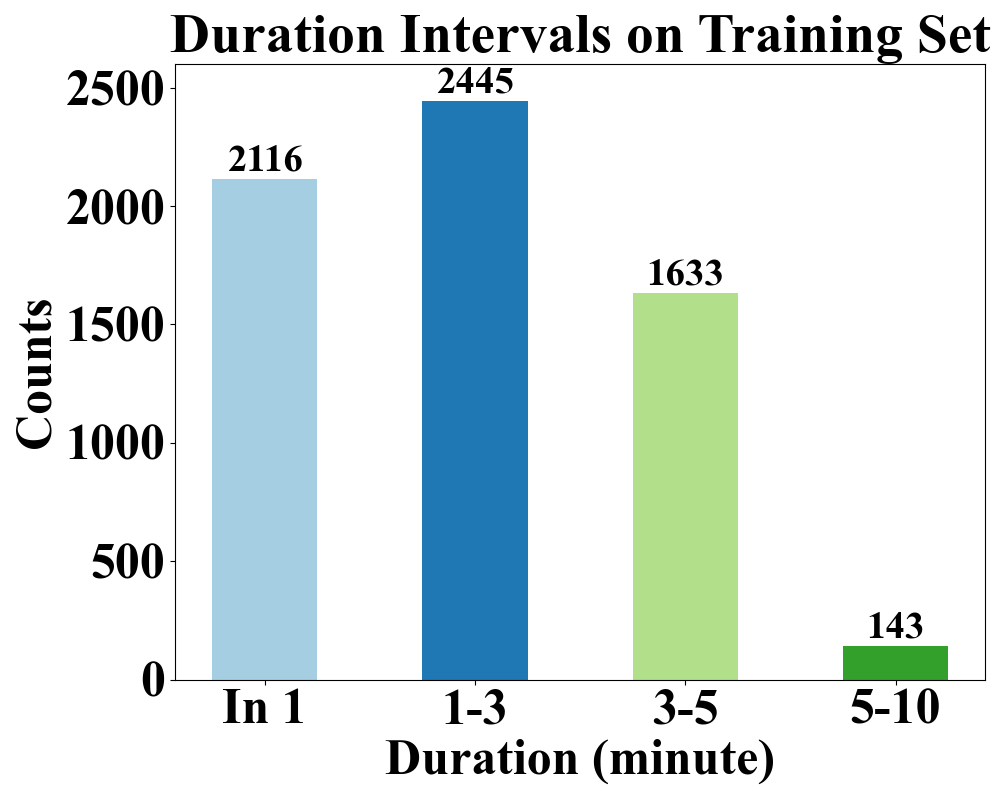}
    \caption{VUDG Training Set}
    \label{fig:vudg_dura_train}
\end{subfigure}
\hspace{30pt}
\begin{subfigure}[t]{0.48\linewidth}
    \includegraphics[width=\linewidth]{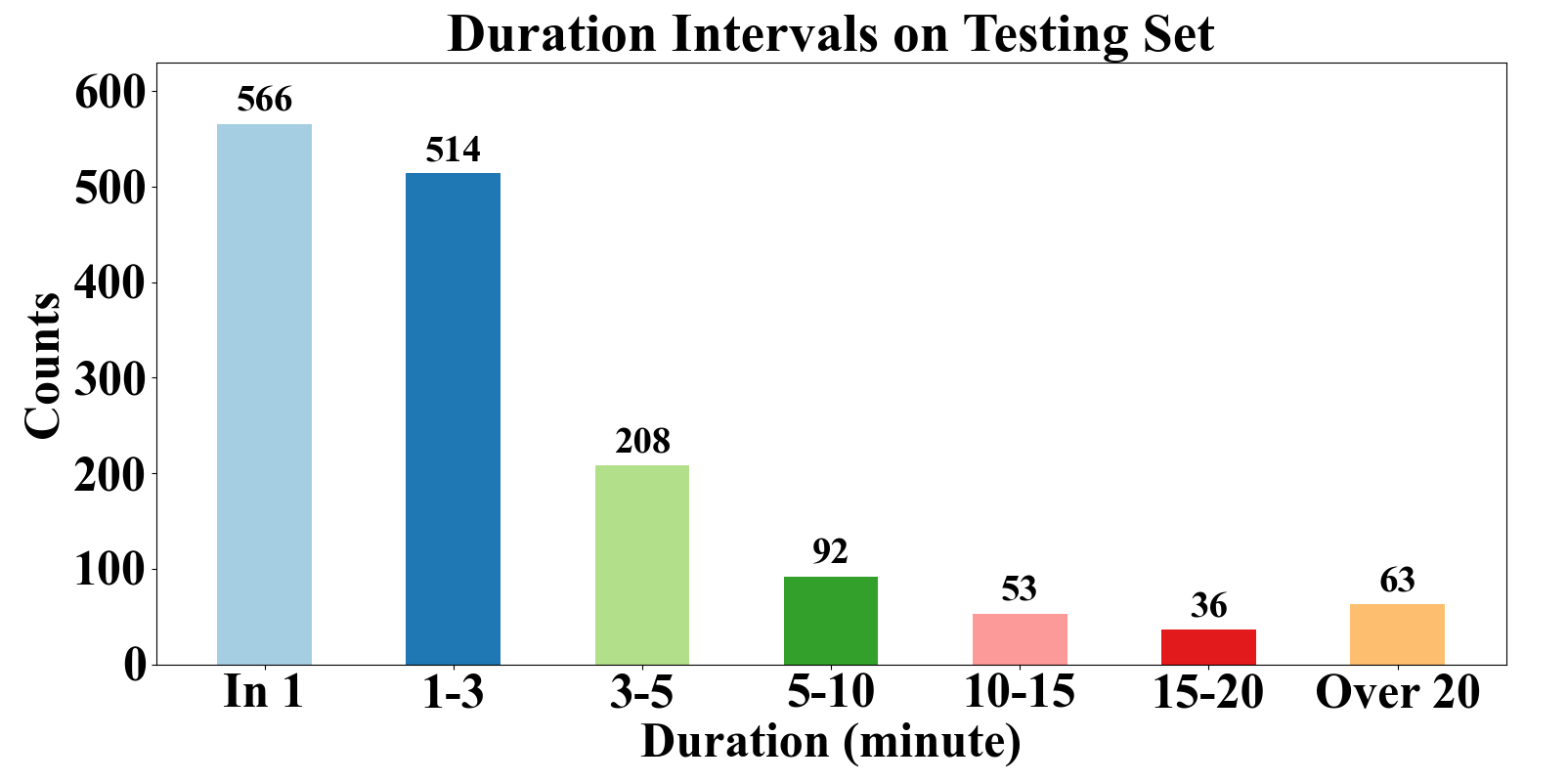}
    \caption{VUDG Testing Set}
    \label{fig:vudg_dura_test}
\end{subfigure}
\caption{Statistics illustrating the distribution of video durations across \textbf{(a)} VUDG training set and \textbf{(b)} VUDG testing set.}
\vspace{-5pt}
\label{fig:vudg_duration_distribution}
\end{figure*}

The training set comprises 6,337 video clips and 31,685 QA pairs. The distribution of videos across domains is illustrated in Figure~\ref{fig:vudg_domain_train}. To reduce memory usage during training, all training videos are limited to a maximum duration of ten minutes. The duration distribution is shown in Figure~\ref{fig:vudg_dura_train}.

The testing set contains 1,532 video clips and 4,703 QA pairs, and the distribution of video numbers in each domain is demonstrated in Figure~\ref{fig:vudg_domain_test}. Compared to the training set, the testing set includes longer videos to better evaluate each model’s ability of handling complex and extended temporal contexts. The duration distribution is shown in Figure~\ref{fig:vudg_dura_test}. 



\subsection{Evaluation Metric}
We adopt two evaluation protocols widely used in domain generalization research: Leave-One-Domain-Out for multiple-domain generalization and Leave-But-One-Domain-Out for single-domain generalization, following prior works~\cite{jo2023poem,chen2023meta}.


For multiple domain generalization, one domain is used as the target domain using its testing set, while the training sets of the remaining $N-1$ domains are treated as source domains for training. The final performance is averaged over all domains, formulated as
\begin{equation}
    \mathrm{Avg}^{m} = \frac{1}{N}\sum_{i=1}^{N} P_i \ ,
\end{equation}
where $P_i$ denotes the accuracy of the $i$-th domain used as the target domain, and $N$ is the number of domains in each setting. 

For single domain generalization, the training set of one domain is used as the source domain for training, and the left $N-1$ domains are treated as target domains using their testing sets. The performance of this domain is calculated as the average of the test accuracy on the left $N-1$ domains, formulated as
\begin{equation}
    \mathrm{Avg}^{s} = \frac{1}{N}\sum_{i=1}^{N}(\frac{1}{N-1}\sum_{j=1,j\neq i}^{N} P_j^i) \ ,
\end{equation}
where $P_j^i$ denotes the accuracy of the $j$-th domain with the $i$-th domain as the source domain. 

For zero-shot generalization, models are directly tested on the full testing sets without any training. For multiple-choice questions, each model receives the video input alongside textual indications, including the question and the candidate options. We then compute the accuracy of a model on each domain and quantify the final performance by averaging the results on all domains.
For open-ended questions, we use DeepSeek-V3 to automatically evaluate the answers. Specifically, we adopt task-specific evaluation protocols to evaluate from two aspects, with a total score of 5 points for each aspect and a maximum score of 10 for each question. For the question category (1)-(3) (Action Recognition, Attribute Identification, and Object Identification), answers are evaluated based on factual accuracy and relevance to the question. For Q4 and Q5 (Temporal Understanding), which emphasize temporal reasoning, answers are assessed based on temporal accuracy and question relevance.
The final score for a question is
\begin{equation}
    \mathrm{Score} = S^{\rm acc} + S^{\rm rel},\quad \mathrm{where} \quad S^{\rm acc},S^{\rm rel} \in[0,5].
\end{equation}
Evaluation prompts for all question types are detailed in the Appendix~\ref{eval_deepseek}.

\section{Experiments}
\vspace{-5pt}
\label{others}

\subsection{Settings}
\vspace{-5pt}
\paragraph{Baseline:}
For domain generalization setting, we evaluate LLM-based and non-LLM-based methods.
HBI~\cite{jin2023video} and EMCL4QA~\cite{jin2022expectationmaximization} are representative VideoQA methods that do not rely on LLMs, while VideoLLaMA2-7B~\cite{damonlpsg2024videollama2} and Qwen2.5VL-3B~\cite{bai2025qwen2} are popular LVLMs using LLMs.
We evaluate their performance 
on the proposed dataset with the multiple domain and single domain generalization settings.
For zero-shot evaluation, we evaluate a total of nine large-scale video understanding models, including Video-ChatGPT-7B~\cite{maaz2024videochatgptdetailedvideounderstanding}, MiniGPT4-Video~\cite{ataallah2024minigpt4}, VideoChat2-7B~\cite{2023videochat}, Video-LLaVA-7B~\cite{lin2023video}, VideoLLaMA2-7B~\cite{damonlpsg2024videollama2}, mPLUG-Owl3-7B-~\cite{ye2025mplugowl}, Video-CCAM-7B~\cite{fei2024videoccamenhancingvideolanguageunderstanding}, VideoLLaMA3-7B~\cite{zhang2025videollama3frontiermultimodal}, Qwen2.5VL-3B and
Qwen2.5VL-7B~\cite{bai2025qwen2}.

\vspace{-5pt}
\paragraph{Implementation Details:}
For training on the VUDG training set, we apply full-parameter fine-tuning to non-LLM-based methods and Low-Rank Adaptation (LoRA)~\cite{hu2022lora} to LVLMs to ensure training efficiency. For LoRA, we set the rank to 128 and the scaling factor to 256. Further training details of each model can be found in Appendix~\ref{implementaion_details}.
For the zero-shot evaluation setting, we set all LVLMs to the official default configuration. Regarding models employing fixed frame sampling, we adopt their default official settings (\emph{e.g.}, Video-LLaMA2 uses 16 frames per video). Regarding models evaluated under fixed FPS settings, we uniformly set the frame rate to 1 FPS to ensure consistency.


\vspace{-5pt}
\subsection{Domain Generalization Results}
\vspace{-5pt}
Table~\ref{tab:lodo} summarizes the performance of four VideoQA methods in multiple domain generalization setting. We test these methods under three types of domain shifts and record the averaged results of each type of domain shift. As shown in Table~\ref{tab:lodo}, we have several findings. Firstly, non-LLM-based methods (HBI~\cite{jin2023video} and EMCL4QA~\cite{jin2022expectationmaximization}) achieve poor performance, indicating their limited generalization ability when exposed to out-of-distribution domains.
Secondly, LLM-based methods show significantly better generalization compared to non-LLM-based methods. Among them, VideoLLaMA2-7B outperforms all others with the highest average accuracy across Visual Style (66.5\%), Viewpoint (66.2\%), and Environmental Condition (68.2\%). Qwen2.5VL-3B shows competitive results but suffers notable degradation under harsh environmental conditions (\emph{e.g.}, only 55.9\% on RA and 55.5\% on SN), suggesting vulnerability to visual noise and degradation.

Despite VideoLLaMA2-7B’s strong performance under multiple DG settings, a clear gap remains compared to models fine-tuned across all domains (Table~\ref{FFTR}). Meanwhile, Qwen2.5-VL-3B shows worse performance on multiple DG fine-tuning than its zero-shot counterpart. These results suggest that current large vision-language models (LVLMs) require more robust training or fine-tuning strategies to enhance domain generalization when adapting to downstream tasks. 


\begin{table}[htbp]
\centering
\caption{Multiple domain generalization test results under different domain shifts, including \textbf{Visual Style}, \textbf{Viewpoint}, and \textbf{Env. Condition} (Environmental Condition). Abbreviations: 
\textbf{CA} (Cartoon),
\textbf{GA} (Game),
\textbf{MO} (Movie/TV),
\textbf{VI} (Virtual),
\textbf{EG} (Egocentric),
\textbf{SU} (Surveillance),
\textbf{SH} (Shaky),
\textbf{FO} (Foggy),
\textbf{NI} (Night),
\textbf{RA} (Rainy),
\textbf{SN} (Snowy), and \textbf{D-Avg} (Domain-wise Average).}

\scalebox{0.65}{\begin{tabular}{l|ccccc|cccc|ccccc|c}
\toprule
\multirow{2}{*}{\textbf{Model}} & 
\multicolumn{5}{c|}{\textbf{Visual Style}} & 
\multicolumn{4}{c|}{\textbf{Viewpoint}} & 
\multicolumn{5}{c|}{\textbf{Env. Condition}} &
\multirow{2}{*}{\textbf{D-Avg}}\\
 & CA & GA & MO & VI & $\mathrm{Avg}^{m}$& EG & SU & SH &$\mathrm{Avg}^{m}$ & FO & NI & RA & SN &$\mathrm{Avg}^{m}$ \\
\midrule
HBI~\cite{jin2023video} & 14.9 & 18.2 & 17.2 & 16.4 &16.7 & 17.4 & 16.7 & 18.5 &17.5 & 17.7 & 18.9 & 16.9 & 17.6 & 17.8 & 17.3\\
EMCL4QA~\cite{jin2022expectationmaximization} &17.7 &18.7 &16.8 &17.7 &17.7 &17.4 &16.7 &19.6&17.9 &19.1 &18.4 &18.8 &18.3 &18.7 & 18.1 \\
Qwen2.5VL-3B~\cite{bai2025qwen2} & \textbf{70.5} & 60.7 & 62.0 & 66.2 &65.0 & 65.6 & 61.2 & 57.7 &61.5 & 61.0 & 55.9 & 59.5 & 55.5 &58.0 & 61.4\\
VideoLLaMA2-7B~\cite{damonlpsg2024videollama2} &61.6  &\textbf{64.4}  &\textbf{68.7}  &\textbf{70.1} &\textbf{66.5} &\textbf{66.1}  &\textbf{62.9}  &\textbf{69.6} &\textbf{66.2} &\textbf{69.6} &\textbf{69.1}  &\textbf{64.9}  &\textbf{69.2}  &\textbf{68.2} &\textbf{66.9}\\
\bottomrule
\end{tabular}}

\label{tab:lodo}
\end{table}

We also evaluate HBI~\cite{jin2023video}, EMCL4QA~\cite{jin2022expectationmaximization}, VideoLLaMA2~\cite{damonlpsg2024videollama2}, and Qwen2.5VL~\cite{bai2025qwen2} in single domain generalization setting under the domain shift caused by the difference of environment condition, \emph{i.e.,} the Environmental Condition split of the VUDG dataset. As shown in Table~\ref{FFTR}, the performance of VideoLLaMA2 degrades 4.7\% compared with that of multiple domain generalization, indicating that single domain generalization poses a greater challenge to these models.
\vspace{-5pt}



\begin{table}[htbp]
\centering
\caption{Single domain generalization results with different domain shift types. Abbreviations: 
\textbf{CA} (Cartoon),
\textbf{GA} (Game),
\textbf{MO} (Movie/TV),
\textbf{VI} (Virtual),
\textbf{EG} (Egocentric),
\textbf{SU} (Surveillance),
\textbf{SH} (Shaky),
\textbf{FO} (Foggy),
\textbf{NI} (Night),
\textbf{RA} (Rainy),
\textbf{SN} (Snowy), and 
\textbf{D-Avg} (Domain-wise Average).
}
\scalebox{0.68}{\begin{tabular}{l|ccccc|cccc|ccccc|c}
\toprule
\multirow{2}{*}{\textbf{Model}} & 
\multicolumn{5}{c|}{\textbf{Visual Style}} & 
\multicolumn{4}{c|}{\textbf{Viewpoint}} & 
\multicolumn{5}{c|}{\textbf{Env. Condition}} &
\multirow{2}{*}{\textbf{D-Avg}}\\
 & CA & GA & MO & VI & ${\mathrm{Avg}^{s}}$& EG & SU & SH &${\mathrm{Avg}^{s}}$ & FO & NI & RA & SN & ${\mathrm{Avg}^{s}}$ \\
\midrule
EMCL4QA~\cite{jin2022expectationmaximization} &18.4 &16.7 &18.5 &17.9 &17.9 &19.0 &17.4 &16.4 &17.6 &18.8  &18.2  &17.1  &16.8&17.7  &17.7\\
HBI\cite{jin2023video} & 18.9 & 16.9 & 17.9 &18.2  &18.0  & 18.4 & 16.9 & 16.8 & 17.4 & 17.9 & 19.3 & 18.6 & 16.8 &18.2 &17.9\\
VideoLLaMA2-7B~\cite{damonlpsg2024videollama2} & 61.7 & 56.8 & 56.2 & 56.8 &57.9  & \textbf{62.4} & \textbf{61.9} & 60.6 &\textbf{62.0}  &  \textbf{68.5} & \textbf{61.4} & \textbf{64.1} & \textbf{60.0} & \textbf{63.5} &60.9 \\
Qwen2.5VL-3B~\cite{bai2025qwen2} & \textbf{63.3} & \textbf{66.5} & \textbf{66.1} & \textbf{64.6} & \textbf{65.1} & 59.8 & 61.7 & \textbf{63.4} & 61.6 & 57.2 & 58.4 & 57.3 & 58.4 & 57.8 & \textbf{61.5 }\\
\bottomrule
\end{tabular}}
\label{FFTR}
\vspace{-5pt}
\end{table}

\vspace{-5pt}
\subsection{Zero-Shot Results}
\vspace{-5pt}

\begin{table}[htbp]
\centering
\caption{
Multi-choice zero-shot test results on VUDG.
Performance across 11 domains. Abbreviations: 
\textbf{CA} (Cartoon),
\textbf{GA} (Game),
\textbf{MO} (Movie/TV),
\textbf{VI} (Virtual),
\textbf{EG} (Egocentric),
\textbf{SU} (Surveillance),
\textbf{SH} (Shaky),
\textbf{FO} (Foggy),
\textbf{NI} (Night),
\textbf{RA} (Rainy),
\textbf{SN} (Snowy), and 
\textbf{D-Avg} (Domain-wise Average).
}
\scalebox{0.7}{\begin{tabular}{l|ccccc|cccc|ccccc|c}
\toprule
\multirow{2}{*}{\textbf{Model}} & 
\multicolumn{5}{c|}{\textbf{Visual Style}} & 
\multicolumn{4}{c|}{\textbf{Viewpoint}} & 
\multicolumn{5}{c|}{\textbf{Env. Condition}} &
\multirow{2}{*}{\textbf{D-Avg}}\\
 & CA & GA & MO & VI & Avg& EG & SU & SH &Avg & FO & NI & RA & SN & Avg \\
\midrule
Video-ChatGPT-7B~\cite{maaz2024videochatgptdetailedvideounderstanding} & 14.1 & 12.5 & 14.4 & 9.7 & 12.7 & 12.9 & 11.6 & 14.1 & 12.9 & 14.5 & 16.3 & 13.8 & 15.8 & 15.1 & 13.6 \\
MiniGPT4-Video~\cite{ataallah2024minigpt4} & 13.6 & 12.9 & 13.9 & 14.2 & 13.7 & 12.7 & 13.2 & 13.8 & 13.2 & 15.9 & 15.7 & 13.0 & 13.0 & 14.4 & 13.8 \\
VideoChat2-7B~\cite{2023videochat} & 16.2 & 9.6 & 14.1 & 10.3 & 12.6 & 14.6 & 13.8 & 13.4 & 13.9 & 16.7 & 15.4 & 17.8 & 11.6 & 15.4 & 14.0 \\
Video-LLaVA-7B~\cite{lin2023video} & 23.3 & 23.1 & 21.2 & 29.9 & 24.4 & 22.3 & 26.1 & 19.5 & 22.6 & 22.6 & 26.0 & 20.1 & 25.0 & 23.4 & 23.5 \\
VideoLLaMA2-7B~\cite{damonlpsg2024videollama2} & 31.5 & 34.0 & 31.7 & 30.4 & 31.9 & 34.6 & 34.5 & 39.6 & 36.2 & 33.4 & 34.7 & 30.5 & 32.2 & 32.7 & 33.4 \\
mPLUG-Owl3-7B~\cite{ye2025mplugowl} & 50.0 & 50.8 & 49.7 & 61.0 & 52.9 & 53.5 & 46.8 & 56.7 & 52.3 & 51.0 & 49.2 & 48.7 & 48.2 & 49.3 & 51.4 \\
Video-CCAM-7B~\cite{fei2024videoccamenhancingvideolanguageunderstanding} & 55.6 & 40.9 & 60.0 & 52.7 & 52.3 & 54.8 & 47.0 & 57.1 & 53.0 & 51.0 & 51.0 & 48.3 & 47.8 & 49.5 & 51.5 \\
VideoLLaMA3-7B~\cite{zhang2025videollama3frontiermultimodal} & 69.7 & \textbf{63.7} & 67.3 & 74.0 & 68.7 & 66.0 & 58.4 & 61.1 & 61.8 & 64.6 & 63.1 & 64.3 & 64.1 & 64.0 & 65.1 \\
Qwen2.5VL-3B~\cite{bai2025qwen2} & \textbf{71.5} & 61.7 & 72.0 & 70.3 & 68.9 & 76.6 & 69.8 & 66.1 & 70.8 & 69.9 & 67.5 & 65.4 & 70.0 & 68.2 & 69.2\\
Qwen2.5VL-7B~\cite{bai2025qwen2} & 71.3 & 61.4 & \textbf{72.3} & \textbf{75.4} & \textbf{70.1} & \textbf{79.8} & \textbf{73.3} & \textbf{68.1} & \textbf{73.7} & \textbf{77.2} & \textbf{69.7} & \textbf{71.0} & \textbf{73.7} & \textbf{72.9} & \textbf{72.1} \\
\bottomrule
\end{tabular}}
\vspace{-5pt}
\label{MZST}
\end{table}

In this section, we report the zero-shot evaluation results on the testing set of VUDG. Domain-specific performance on multiple-choice QA pairs is presented in Table~\ref{MZST}, while the scoring outcomes for open-ended questions are summarized in Table~\ref{OZST}. For both types, we provide detailed results as well as domain-wise averages (D-Avg) across 11 domains. Table~\ref{MTQC} presents the accuracy of multiple-choice questions categorized by question type. 

\paragraph{Multiple-Choice QA:}
As shown in Table 2, Qwen2.5VL-7B achieves the highest average accuracy (72.1\%) across 11 visual domains, demonstrating strong generalization. Notably, models like Video-LLaMA3-7B show moderate performance but struggle on viewpoint domain shifts such as Surveillance (SU) and Shaky (SH) scenes. Earlier models like Video-ChatGPT-7B and MiniGPT4-Video exhibit much lower performance overall, suggesting limited robustness to various distribution shifts.

\begin{table}[htbp]
\centering
\caption{
Open-ended zero-shot test results on VUDG.
Performance across 11 visual domains. Abbreviations: 
\textbf{CA} (Cartoon),
\textbf{GA} (Game),
\textbf{MO} (Movie/TV),
\textbf{VI} (Virtual),
\textbf{EG} (Egocentric),
\textbf{SU} (Surveillance),
\textbf{SH} (Shaky),
\textbf{FO} (Foggy),
\textbf{NI} (Night),
\textbf{RA} (Rainy),
\textbf{SN} (Snowy), and 
\textbf{D-Avg} (Domain-wise Average).
}
\scalebox{0.7}{\begin{tabular}{l|ccccc|cccc|ccccc|c}
\toprule
\multirow{2}{*}{\textbf{Model}} & 
\multicolumn{5}{c|}{\textbf{Visual Style}} & 
\multicolumn{4}{c|}{\textbf{Viewpoint}} & 
\multicolumn{5}{c|}{\textbf{Env. Condition}} &
\multirow{2}{*}{\textbf{D-Avg}}\\
 & CA & GA & MO & VI & Avg & EG & SU & SH & Avg & FO & NI & RA & SN & Avg \\
\midrule
MiniGPT4-Video~\cite{ataallah2024minigpt4} & 3.77 & 3.40 & 4.47 & 4.42 & 4.02 & 4.86 & 3.91 & 4.79 & 4.52 & 4.69 & 4.56 & 4.03 & 4.61 & 4.47 & 4.32\\
Video-ChatGPT-7B~\cite{maaz2024videochatgptdetailedvideounderstanding} & 5.34 & 5.10 & 5.63 & 5.63 & 5.43 & 5.88 & 5.36 & 5.87 & 5.70 & 5.70 & 5.81 & 5.31 & 5.87 & 5.67 & 5.59 \\
VideoChat2-7B~\cite{2023videochat} & 5.22 & 5.33 & 5.71 & 5.82 & 5.52 & 6.12 & 5.44 & 5.98 & 5.85 & 5.77 & 5.96 & 5.17 & 5.88 & 5.70 & 5.67 \\
VideoLLaMA3-7B~\cite{zhang2025videollama3frontiermultimodal} & 5.57 & 5.35 & 5.80 & 5.91 & 5.66 & 5.77 & 5.64 & 5.93 & 5.78 & 5.61 & 5.75 & 5.50 & 5.83 & 5.67 & 5.70 \\
Video-LLaVA-7B~\cite{lin2023video} & 5.39 & 5.22 & 5.81 & 6.20 & 5.66 & 6.32 & 5.66 & 6.11 & 6.03 & 5.80 & 5.99 & 5.42 & 6.03 & 5.81 & 5.81 \\
VideoLLaMA2-7B~\cite{damonlpsg2024videollama2} & 5.80 & 5.91 & 6.32 & 6.68 & 6.18 & 6.88 & 6.05 & 6.65 & 6.53 & 6.46 & 6.55 & 6.12 & 6.45 & 6.40 & 6.35 \\
mPLUG-Owl3-7B~\cite{ye2025mplugowl} & 6.07 & 6.10 & 6.44 & 6.98 & 6.40 & 7.32 & 6.07 & 6.92 & 6.77 & 6.75 & 6.84 & 6.33 & 6.76 & 6.67 & 6.60 \\
Qwen2.5VL-7B~\cite{bai2025qwen2} & \textbf{6.53} & 6.23 & \textbf{6.85} & 7.02 & 6.66 & \textbf{7.46} & 6.44 & 6.99 & 6.96 & \textbf{6.98} & 6.66 & \textbf{6.66} & \textbf{6.87} & 6.79 & 6.79 \\
Video-CCAM-7B~\cite{fei2024videoccamenhancingvideolanguageunderstanding} & 6.51 & \textbf{6.46} & 6.75 & \textbf{7.22} & \textbf{6.74} & 7.43 & \textbf{6.45} & \textbf{7.15} & \textbf{7.01} & 6.93 & \textbf{6.89} & 6.55 & \textbf{6.87} & \textbf{6.81} & \textbf{6.84} \\
\bottomrule
\end{tabular}}
\label{OZST}
\end{table}

\paragraph{Open-End QA:}
As shown in Table~\ref{OZST}, Video-CCAM again achieves the highest domain average (6.84), followed closely by Qwen2.5VL-7B and mPLUG-Owl3-7B. Compared to the multiple-choice setting, performance gaps among models narrow in the open-ended QA setting, suggesting increased difficulty in generating free-form answers under domain shifts. The results also reveal substantial variation across domains, with performance dropping obviously in challenging visual conditions like Cartoon, Game, and Surveillance. 

\begin{table}[htbp]
\centering
\caption{
Various question categories of multiple-choice test results on VUDG.
\textbf{Q1} (Action Recognition), 
\textbf{Q2} (Attribute Identification), 
\textbf{Q3} (Object Identification), 
\textbf{Q4 \& Q5} (Temporal Understanding).
}
\scalebox{0.75}{\begin{tabular}{l|c|c|c|c|c}
\toprule
\textbf{Model} & \textbf{Q1} & \textbf{Q2} & \textbf{Q3} & \textbf{Q4 \& Q5} & \textbf{Overall}\\
\midrule
Video-ChatGPT-7B~\cite{maaz2024videochatgptdetailedvideounderstanding} & 11.6 & 13.4 & 12.7 & 14.7 & 13.6 \\
VideoChat2-7B~\cite{2023videochat} & 15.3 & 13.6 & 12.4 & 13.7 & 13.7\\
MiniGPT4-Video~\cite{ataallah2024minigpt4} & 14.4 & 11.8 & 14.9 & 14.1 & 13.8 \\
Video-LLaVA-7B~\cite{lin2023video} & 30.1 & 28.2 & 22.1 & 21.0 & 24.1\\
VideoLLaMA2-7B~\cite{damonlpsg2024videollama2} & 35.3 & 36.2 & 36.3 & 30.2 & 33.3 \\
Video-CCAM-7B~\cite{fei2024videoccamenhancingvideolanguageunderstanding} & 53.8 & 55.0 & 58.0 & 47.1 & 51.5 \\
mPLUG-Owl3-7B~\cite{ye2025mplugowl} & 55.2 & 53.8 & 58.9 & 46.8 & 51.6 \\
VideoLLaMA3-7B~\cite{zhang2025videollama3frontiermultimodal} & 64.3 & 75.2 & 73.7 & 59.0 & 65.4 \\
Qwen2.5VL-3B~\cite{bai2025qwen2} & 71.7 & 73.4 & 77.0 & 64.9 & 69.7 \\
Qwen2.5VL-7B~\cite{bai2025qwen2} & \textbf{73.7} & \textbf{77.3} & \textbf{80.8} & \textbf{67.7} & \textbf{72.7} \\
\bottomrule
\end{tabular}}
\label{MTQC}
\end{table}

\paragraph{Performance Across Question Categories:}
To further understand model behavior, Table~\ref{MTQC} breaks down performance by question category. Qwen2.5VL-7B achieves the best performance across all categories, with particularly strong results in Object Identification (Q3) and Temporal Understanding (Q4 \& Q5). In addition, the performance of most models in action recognition, attribute identification, and object identification is generally better than that in temporal understanding. This pattern suggests that current LVLMs are better at static, appearance-based reasoning than dynamic temporal understanding, revealing a challenge for video-based generalization.

\vspace{-5pt}
\subsection{Visualization}
\vspace{-5pt}

\begin{figure}
    \centering
    \includegraphics[width=0.9 \linewidth]{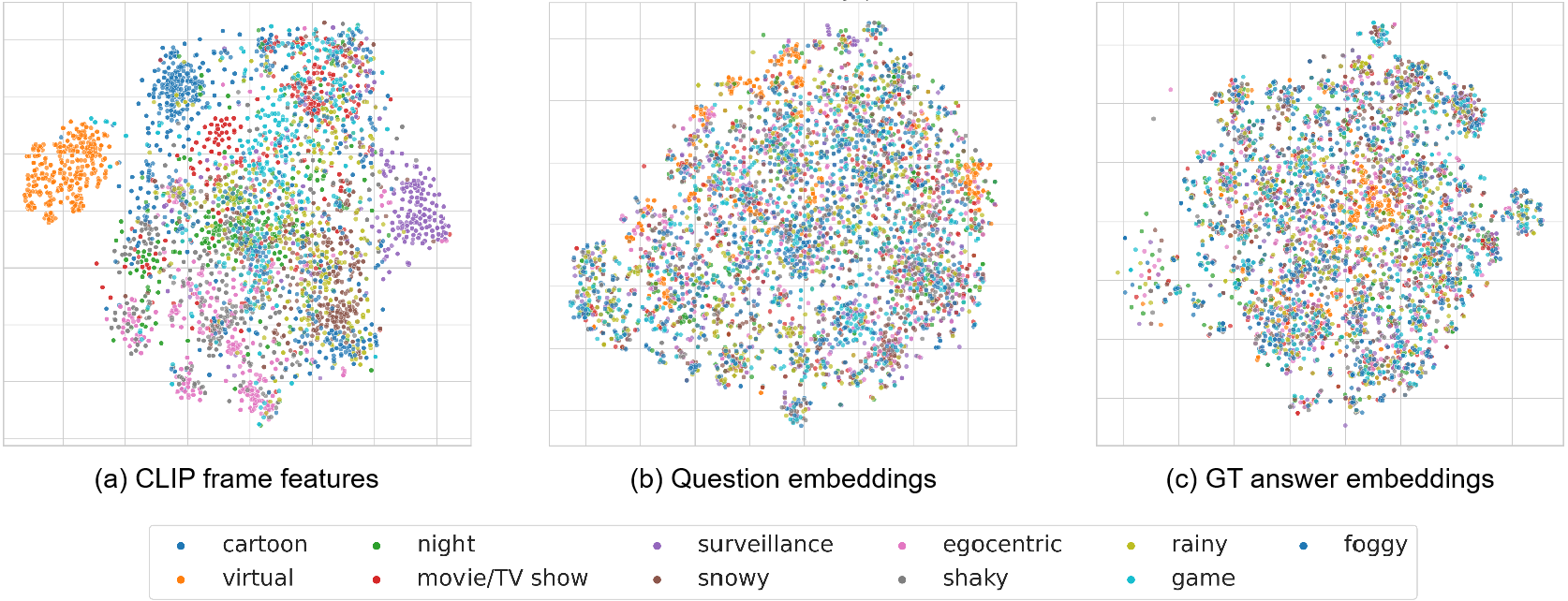}
    \caption{The t-SNE visualization of frame features, question embeddings, and ground-truth answer embeddings across different domains.}
    \label{fig:t-SNE}
\end{figure}
We visualize the embeddings of video frames, questions, and ground-truth answers using CLIP-B/16~\cite{dosovitskiy2021an} for image encoding and BGE-V1.5~\cite{10.1145/3626772.3657878} for text encoding. As shown in Fig.~\ref{fig:t-SNE}, the CLIP frame features (a) exhibit relatively clear clustering by domain, indicating significant visual differences across domains and validating the challenge of domain shifts in VUDG. In contrast, the distributions of question embeddings (b) and answer embeddings (c) are more uniformly mixed, suggesting that the semantic content remains consistent across domains. This supports our design principle of preserving cross-domain semantic similarity while introducing realistic visual shifts.

\vspace{-5pt}
\section{Conclusions}
\vspace{-5pt}

We present VUDG, a novel dataset for evaluating domain generalization (DG) in video understanding while maintaining semantic similarity across 11 diverse domains, enabling fair and challenging evaluation. 
To construct high-quality QA pairs, we develop a progressive multi-expert annotation pipeline that benefits from multiple large models together with human expert refinement.
Through extensive experiments, we observe that current models struggle to address domain shifts when adapted to downstream tasks, resulting in suboptimal generalization. As for zero-shot evaluation, the performance of different LVLMs varies greatly, while even state-of-the-art LVLMs exhibit inconsistent performance between domains. 
Additionally, models fine-tuned on the complete training set of VUDG achieve substantial gains, demonstrating the effectiveness of our training set in improving generalization ability. 
We believe VUDG provides a valuable foundation for advancing generalizable video understanding. However, our dataset currently only focuses on visual domain shifts. In the future, we plan to explore textual domain shifts and incorporate additional modalities such as audio to enable more comprehensive multimodal generalization.

\bibliographystyle{unsrt}


\bibliography{ref.bib}

\newpage
\appendix

\section{Domain Definitions}
The definitions of all domains are shown in Table~\ref{tab:domain_descriptions}.

\begin{table}[htbp]
\centering
\caption{Domain definitions grouped by distribution shift types, along with descriptions.}
\scalebox{0.8}{\begin{tabular}{lll}
\toprule
\textbf{Shift Type} & \textbf{Domain} & \textbf{Description} \\
\midrule
\multirow{4}{*}{Visual Style} 
& Cartoon     & Stylized animation with exaggerated motion and simplified texture \\
& Game        & Scenes from video games, often rendered in real-time \\
& Movie/TV   & Professionally shot narrative content with cinematic framing \\
& Virtual     & Fully simulated environments, often from virtual production \\
\midrule
\multirow{3}{*}{Viewpoint} 
& Egocentric  & First-person perspective, often from head-mounted cameras \\
& Surveillance & Static, wide-angle views from mounted cameras in public/private spaces \\
& Shaky       & Handheld, unstable footage with dynamic camera motion \\
\midrule
\multirow{4}{*}{Environmental Condition} 
& Foggy       & Reduced visibility due to simulated or natural fog \\
& Night       & Low-light or nighttime scenes, often under artificial lighting \\
& Rainy       & Outdoor scenes with visible rain, wet surfaces, and overcast skies \\
& Snowy       & Scenes with snowfall, snow-covered ground, and diffused light \\
\bottomrule
\end{tabular}}
\label{tab:domain_descriptions}
\end{table}

\section{Data Collection}
\label{prompt_data_collection}
\begin{figure}[htbp]

    \centering
    \includegraphics[width=0.9\textwidth]{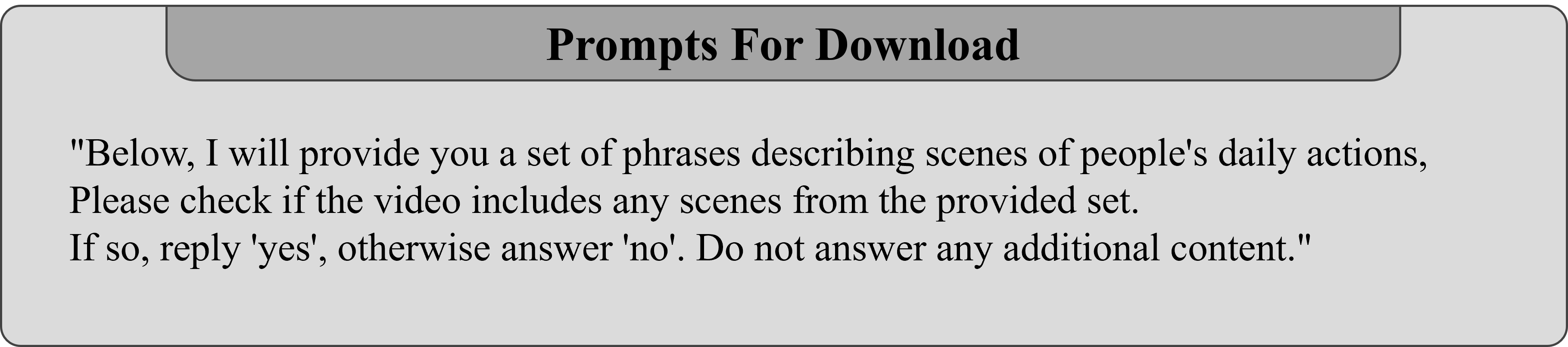}
    \caption{Prompts for download.}
    \label{fig:dld}
\end{figure}

We present the prompt for Qwen2.5-VL-3B used to select videos that belongs to the specific activity list in Figure~\ref{fig:dld} and the types of daily actions that the VUDG dataset focuses on are shown in Table~\ref{DALD}.

The list of human daily activity scenes containing 37 items that the VUDG dataset focuses on are detailed in Table~\ref{DALD}.

\begin{longtable}[htbp]{p{0.25\textwidth} p{0.7\textwidth}}
\label{DALD} \\
\caption{Human Daily Activity Scenes List with Descriptions} \\
\toprule
\textbf{Activity} & \textbf{Description} \\
\midrule
\endfirsthead
\toprule
\textbf{Activity} & \textbf{Description} \\
\midrule
\endhead
Sitting & Sitting on a couch or chair \\
Using Laptop & Using a laptop or other electronic devices (including typing on a keyboard) \\
Writing/Drawing & Writing or drawing in a notebook (including painting) \\
Cooking & Cooking or preparing food (chopping, cutting, mixing, stirring) \\
Cleaning Dishes & Cleaning dishes or household items; organizing spaces \\
Gardening & Gardening or plant care (watering, tending, pruning) \\
Exercising & Exercising or stretching (yoga, home workouts, outdoor) \\
Hair Styling & Brushing or styling hair (combing, etc.) \\
Reading & Reading books or documents \\
Hygiene & Washing hands or doing hygiene routines (brushing teeth) \\
Meeting & Attending meetings or presentations (online or in-person) \\
Taking Notes & Taking notes during a meeting \\
Drinking & Drinking water, coffee, tea, or other beverages \\
Preparing Meals & Preparing a meal or snack (lunch, dinner) \\
Using Phone & Using a phone or tablet (checking notifications) \\
Walking Outdoors & Walking in a park or natural area \\
Fixing Items & Adjusting or fixing household items \\
Feeding Pets & Feeding a pet \\
Art/Crafting & Engaging in art or crafting activities \\
Device Setup & Setting up or adjusting electronic devices \\
Checking Calendar & Checking or updating a calendar or planner \\
Watching Screens & Watching television or screens (incl. gaming) \\
Resting & Resting or lying down \\
Tool Preparation & Preparing tools or items for a task \\
Organizing & Cleaning or organizing a room (tidying clutter) \\
Laundry & Sorting laundry or folding clothes \\
Taking a Break & Taking a break from a task or activity \\
Conversation & Engaging in discussions or conversations \\
Shopping & Shopping or browsing items \\
Driving & Driving a car \\
Eating & Eating a meal \\
Playing Toys & Playing with a toy \\
Makeup & Applying makeup \\
Dancing & Dancing \\
Biking & Riding a bicycle \\
Music & Playing a musical instrument \\
Commuting & Commuting via public transport \\
\bottomrule
\end{longtable}

\section{Implementation Details}
\label{implementaion_details}
All the experiments are conducted on 8 NVIDIA RTX 4090D GPUs, each with 24GB of graphics memory.
\subsection{HBI}




We train HBI for 5 epochs with a global batch size of 32. On the model side, we cap captions at 32 words and videos at 12 frames, employ a 2D linear patch mode, set the slice‐frame position to 2, keep all layers trainable (no freezing), and enable ``loose" interaction modeling. Optimization is carried out with a base learning rate of 1e-4, a higher coefficient learning rate of 1e-3 for specialized submodules, and weighted loss terms of 2 for the KL divergence and 1 for the symmetric KL (SKL) divergence.
\subsection{EMCL4QA}
 We integrate the EMCL module into a CLIP‑based video–question–answering backbone and train for 5 epochs with a global batch size of 128, clipping videos to 12 frames and questions to 32 tokens. During each forward pass, EMCL performs $T=9$ routing iterations over $K=32$ subspaces using a Gaussian kernel ($\sigma=1$) and updates its bases via moving‐average momentum $\alpha=0.9$, then fuses reconstructed and original features with scale factor $\beta=0.5$. We optimize end‑to‑end with Adam and a 10\% linear warmup, applying a learning rate of 1e-7 to the CLIP encoders and 1e-4 to the EMCL module and QA head, training against an InfoNCE‐based cross‐entropy loss with temperature $\tau=0.01$.

\subsection{VideoLLaMA2}
We fine-tune VideoLLaMA2 with a LoRA rank of $r=128$ and an alpha of $256$, inserting STC Connector modules as multimodal projectors learned at 2e-5. We train with a global batch size of $64$ (per device 1, gradient accumulation 8), optimize via AdamW at 2e-5 with no weight decay and a 0.03 warmup ratio, and apply a cosine annealing schedule throughout. To enable efficient large-scale training, we leverage DeepSpeed with ZeRO Stage 3 optimization, which allows for memory-efficient distributed training without compromising performance.

\subsection{Qwen}

We finetune Qwen2.5VL-3B using LoRA with a rank of 64 and a LoRA alpha of 64, targeting key projection modules (\textit{q\_proj}, \textit{k\_proj}, \textit{v\_proj}, \textit{o\_proj}) within the model architecture. We adopt a global batch size of 32, achieved by setting a per-device batch size of 4 and gradient accumulation steps of 4. Optimization is performed using the AdamW optimizer with a learning rate of 2e-7, no weight decay, and a warmup ratio of 0.03. A cosine learning rate scheduler is employed throughout training. To enable efficient large-scale training, we leverage DeepSpeed with ZeRO Stage 3 optimization, which allows for memory-efficient distributed training without compromising performance.

\section{Societal Impacts}
Our work aims to improve the robustness and generalization of video-language models in real-world scenarios by introducing a dataset focused on domain generalization. This has potential positive societal impacts in enhancing the reliability of AI systems used in safety-critical or diverse environments, such as surveillance, autonomous vehicles, or assistive technologies, where domain shifts are inevitable. 

Similar to many existing datasets, VUDG includes some publicly available videos collected from online platforms. While we take care to avoid personally identifiable or sensitive content, the use of web-sourced data may still raise concerns regarding content ownership or individual rights. To mitigate such risks, our dataset is released strictly for non-commercial, academic research purposes only. We encourage responsible use and further discussion on ethical data curation and model deployment practices in this field.

\section{Prompts for generation and evaluation}

\subsection{Prompts for open-end QA pairs generation}
\label{prompt_openend}
\begin{figure}[htbp]

    \centering
    \includegraphics[width=1\textwidth]{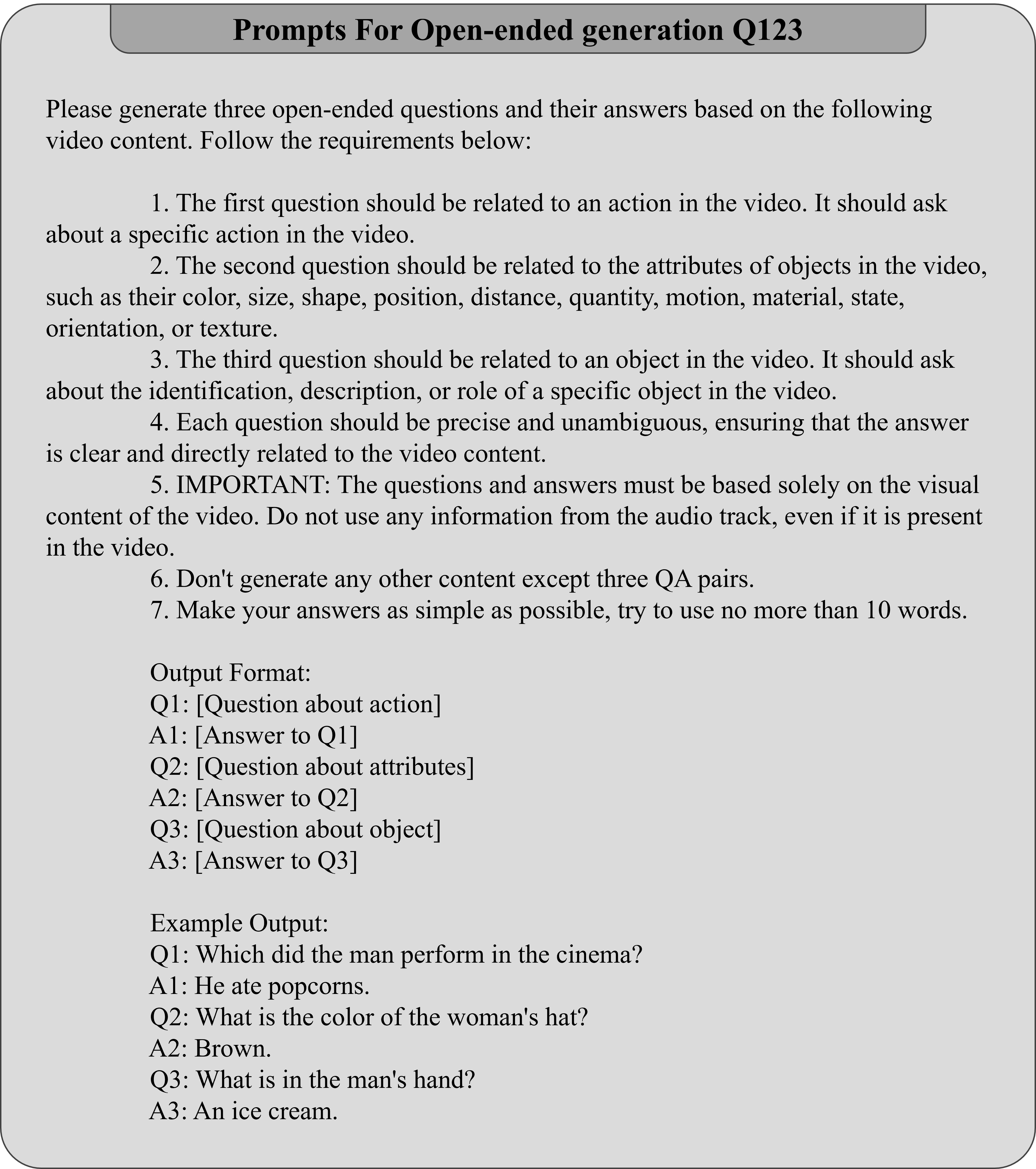}
    \caption{Prompts for generating open-ended answers for Q1, Q2 and Q3.}
    \label{fig:open123}
\end{figure}

\begin{figure}[htbp]

    \centering
    \includegraphics[width=1\textwidth]{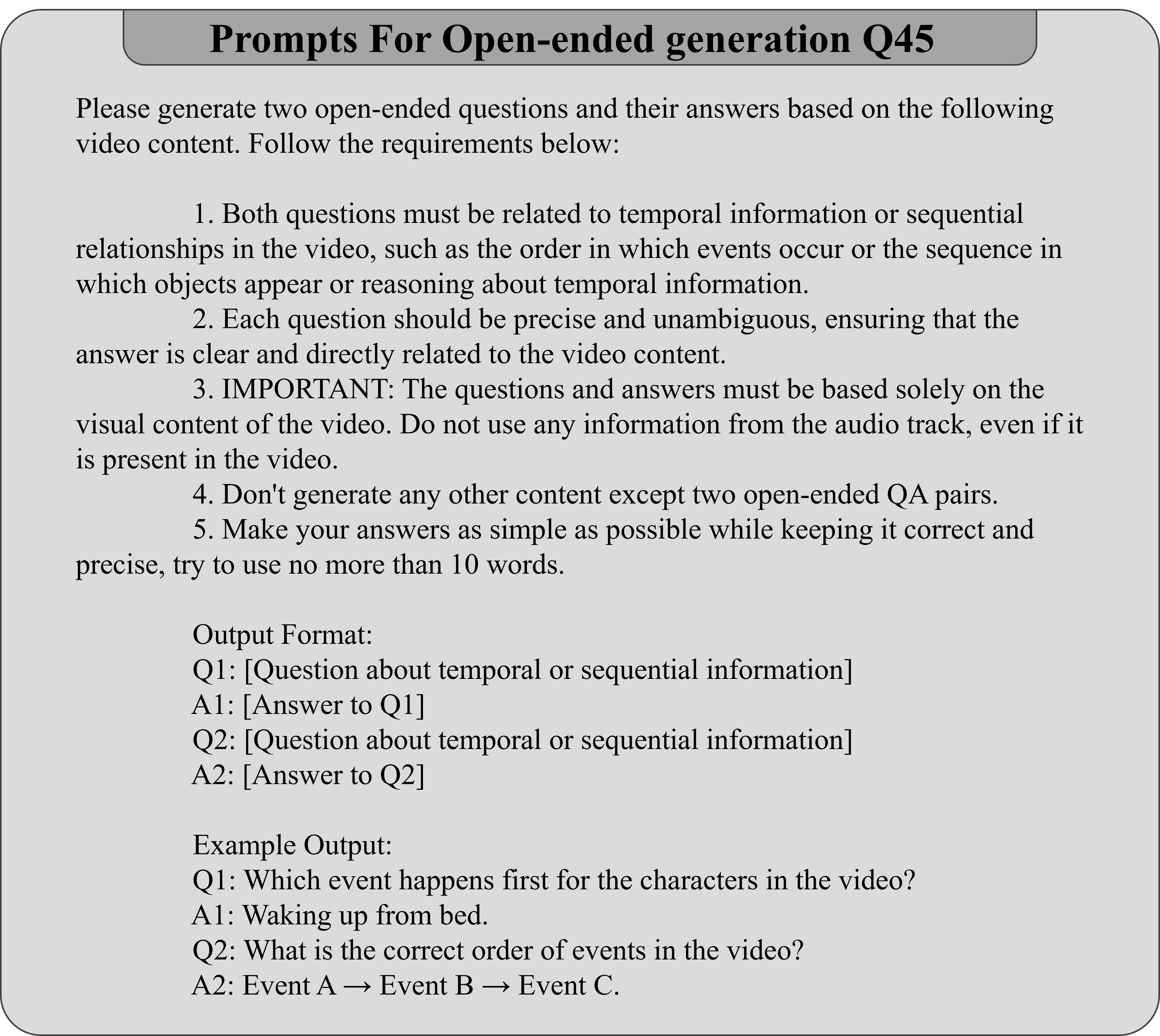}
    \caption{Prompts for generating open-ended answers for Q4 and Q5.}
    \label{fig:open45}
\end{figure}

Prompts for open-ended QA pairs generation are shown in Figure~\ref{fig:open123} and Figure~\ref{fig:open45}. 

\subsection{Prompts for multiple-choice options generation}
\label{prompt_multichoice}
\begin{figure}[htbp]

    \centering
    \includegraphics[width=1\textwidth]{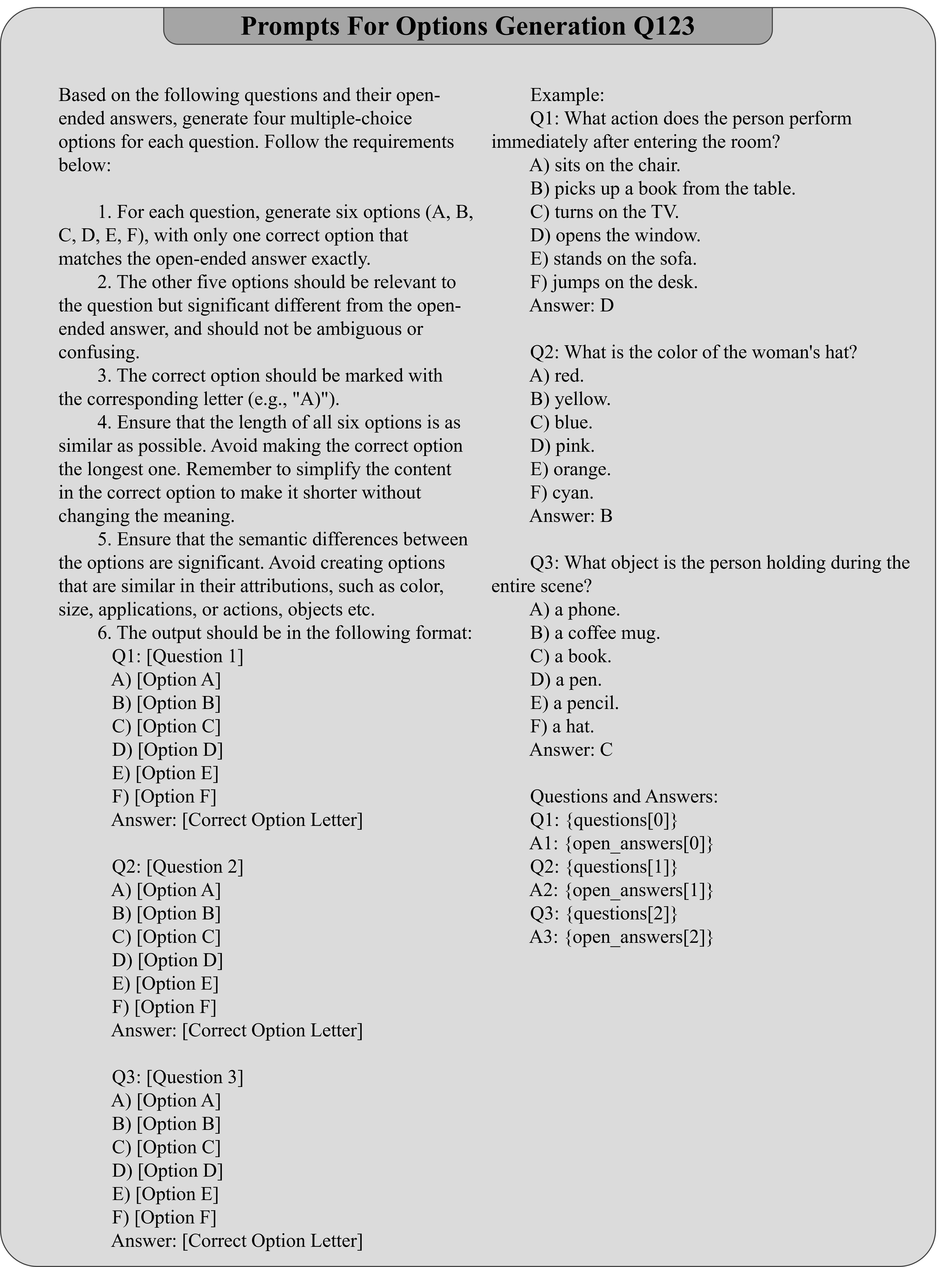}
    \caption{Prompts for generating options for Q1, Q2 and Q3.}
    \label{fig:opt123}
\end{figure}

\begin{figure}[htbp]

    \centering
    \includegraphics[width=1\textwidth]{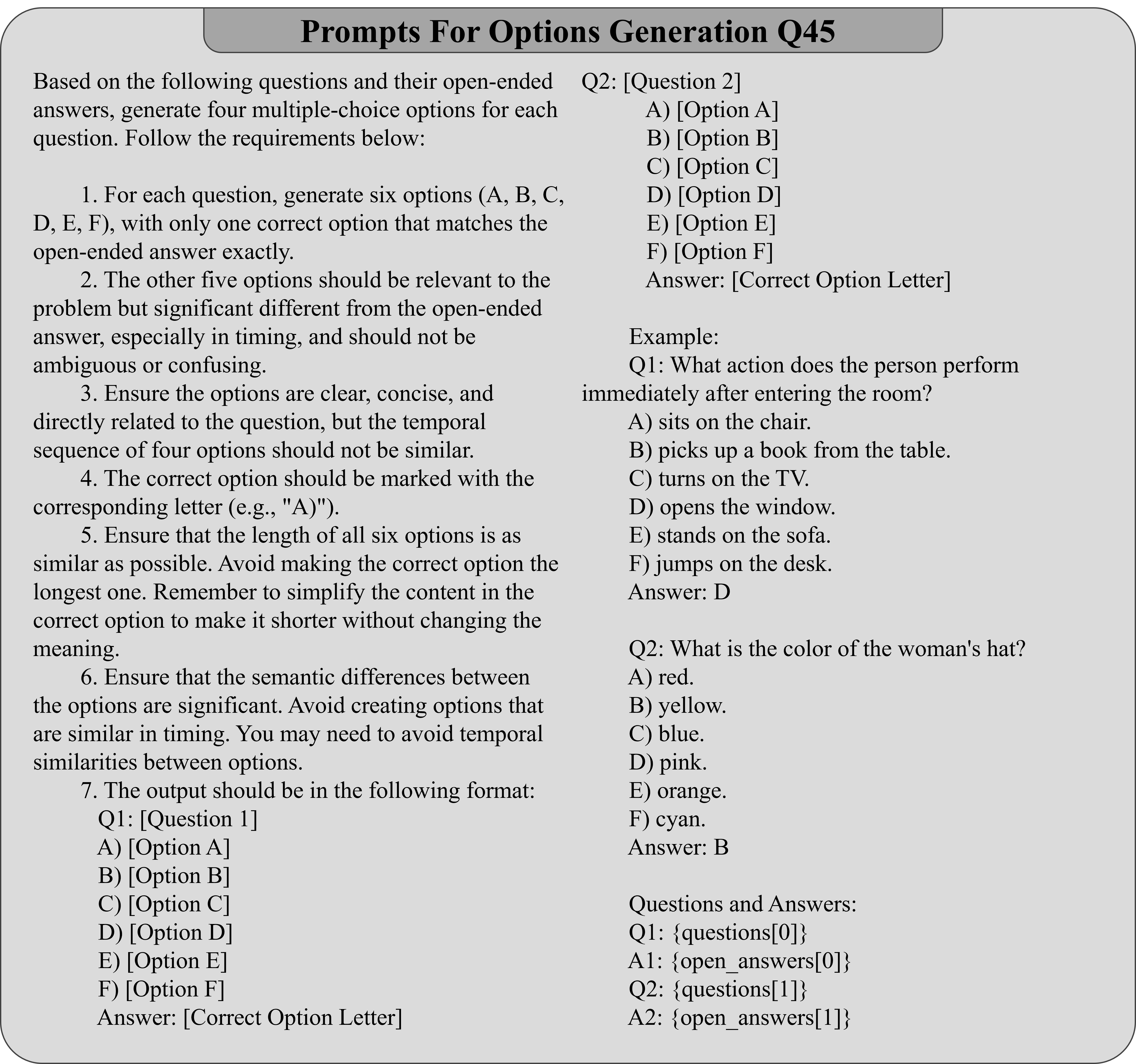}
    \caption{Prompts for generating options for Q4 and Q5.}
    \label{fig:opt45}
\end{figure}

Prompts for multiple-choice options generation are shown in Figure~\ref{fig:opt123} and Figure~\ref{fig:opt45}. 

\subsection{Prompt for reviewing QA pairs}
\label{QA_review}
\begin{figure}[htbp]
    
    \includegraphics[width=1\linewidth]{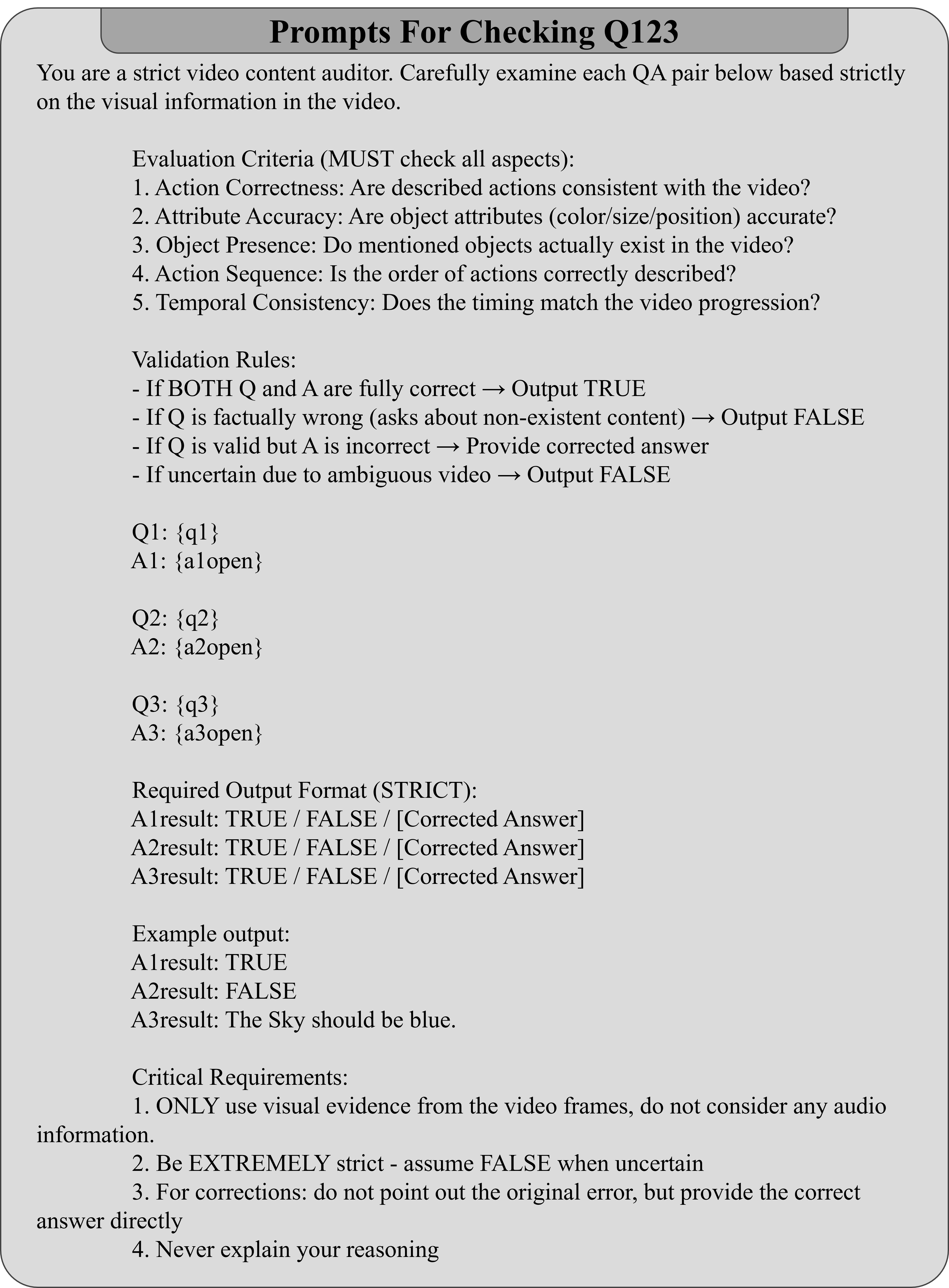}
    \caption{Prompts for Checking Q1, Q2 and Q3. }
    \label{fig:Check123}
\end{figure}

\begin{figure}[htbp]
    
    \includegraphics[width=1\linewidth]{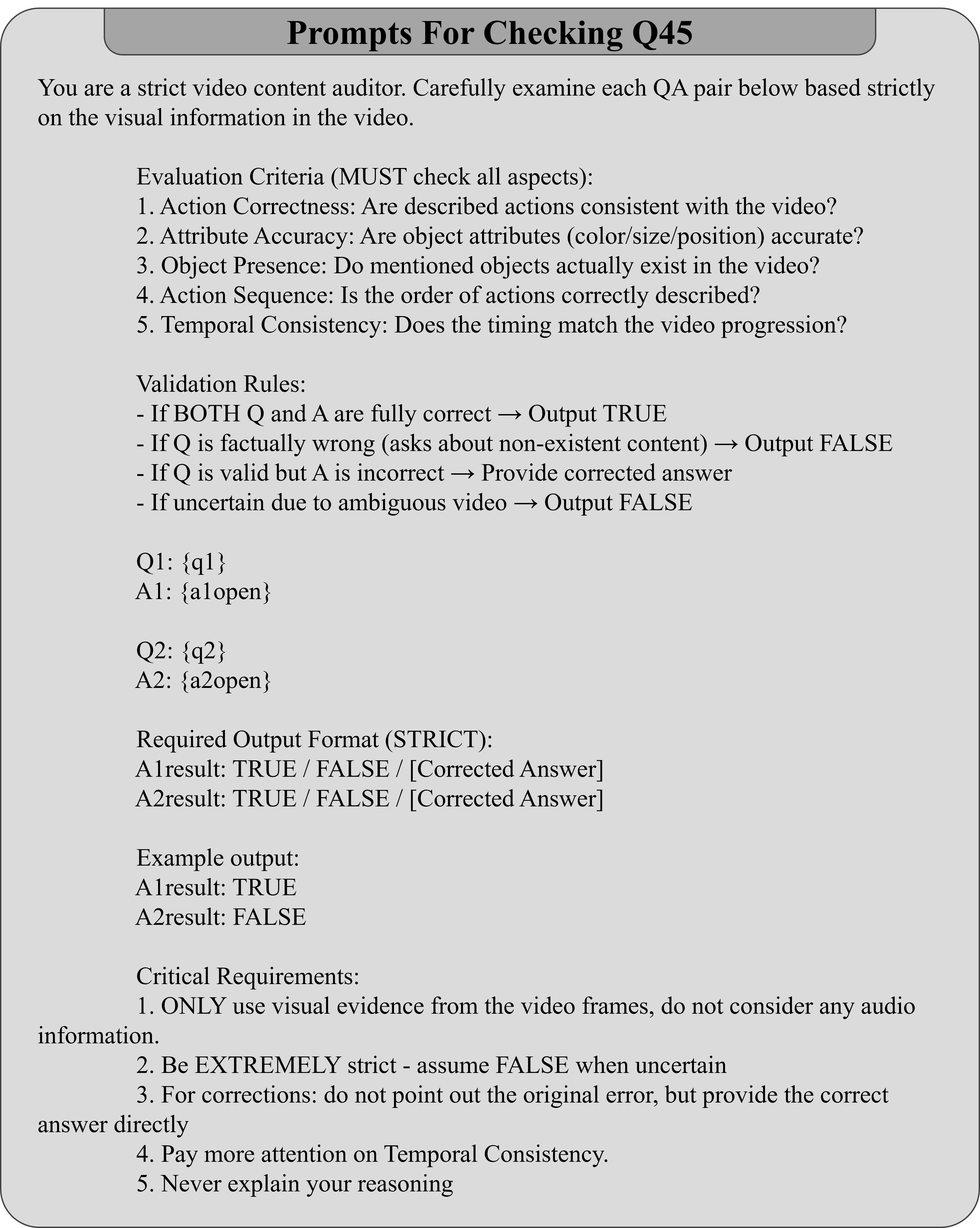}
    \caption{Prompts for Checking Q4 and Q5.}
    \label{fig:Check45}
\end{figure}

Prompts for reviewing QA pairs using Gemini-2.5-Pro are shown in Figure~\ref{fig:Check123} and Figure~\ref{fig:Check45}.

\subsection{Prompts for Open-ended Evaluation}
\label{eval_deepseek}
\begin{figure}[htbp]
    
    \includegraphics[width=1\linewidth]{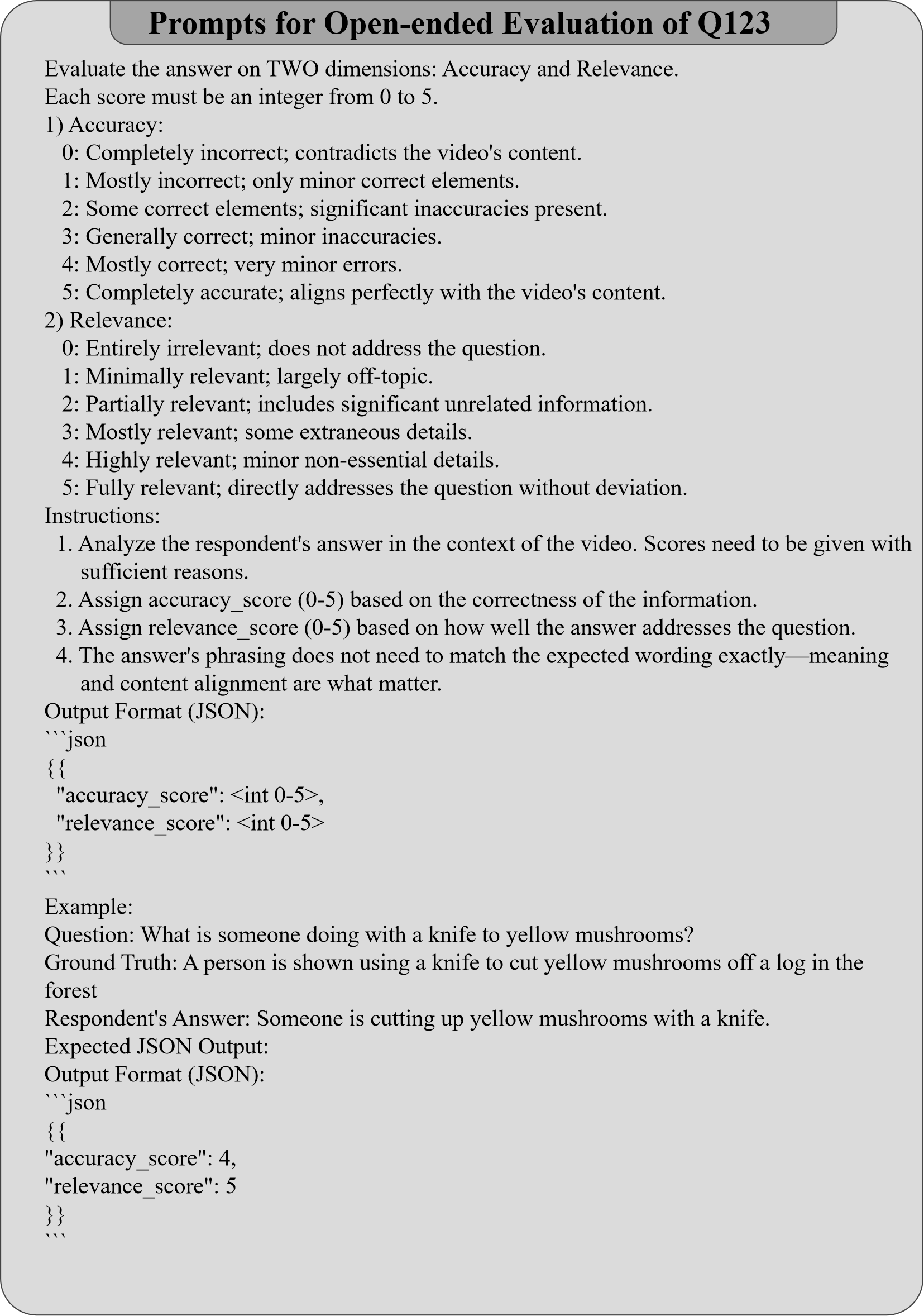}
    \caption{Prompts for Open-ended Evaluation of Q1 to Q3(Action Recognition, Attribute Identification, and Object Identification)}
    \label{fig:evaluate_openend_q123}
\end{figure}

\begin{figure}[htbp]
    
    \includegraphics[width=1\linewidth]{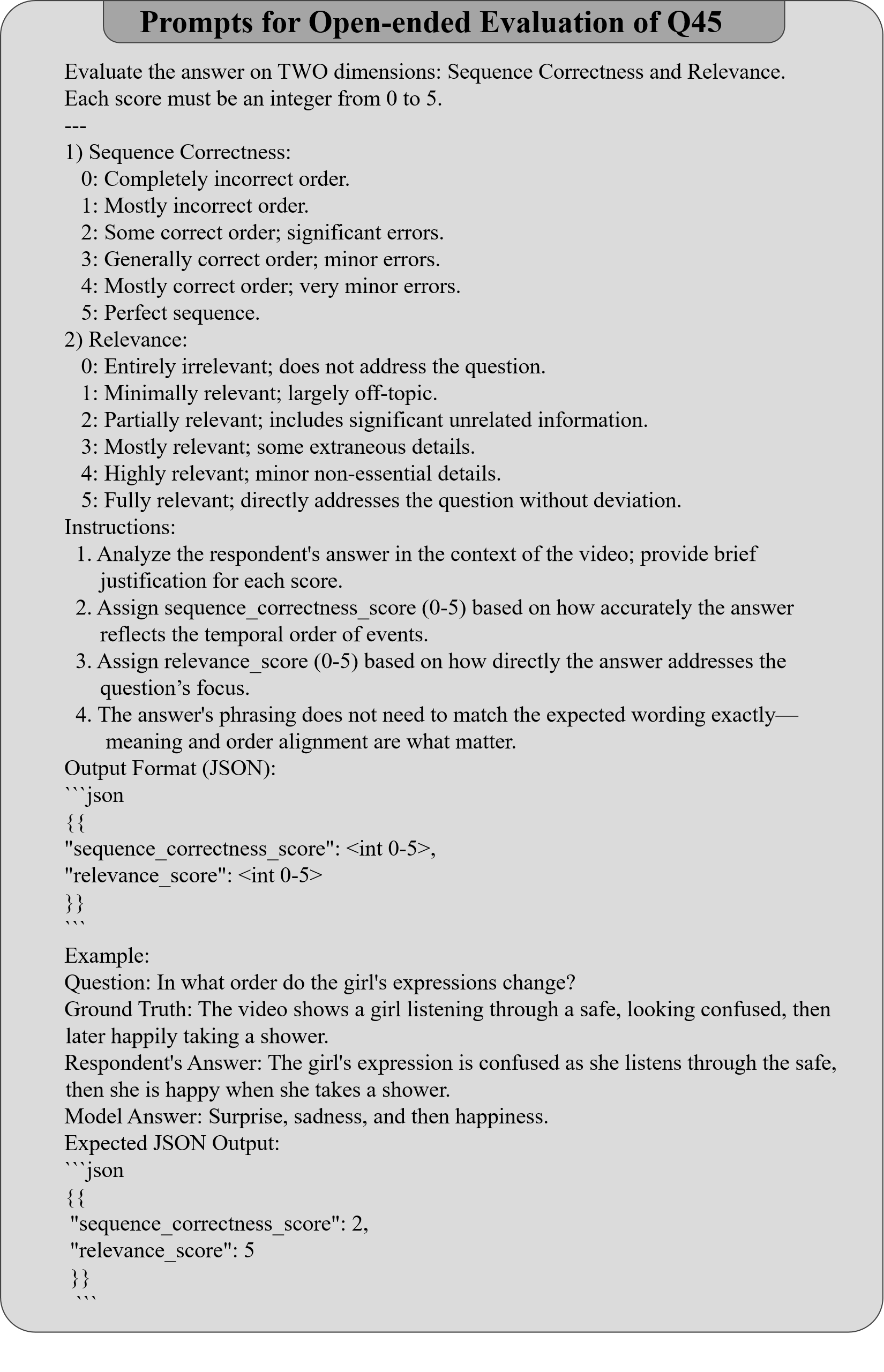}
    \caption{Prompts for Open-ended Evaluation of Q4 and Q5(Temporal Understanding)}
    \label{fig:evaluate_openend_q45}
\end{figure}

Prompts for open-ended evaluation using DeepSeek-V3 are shown in Figure~\ref{fig:evaluate_openend_q123} and Figure~\ref{fig:evaluate_openend_q45}.


\end{document}